\pdfoutput=1

\documentclass[11pt]{article}

\usepackage[]{acl}

\usepackage{times}
\usepackage{latexsym}

\usepackage[T1]{fontenc}

\usepackage[utf8]{inputenc}

\usepackage{microtype}

\usepackage{inconsolata}

\usepackage{graphicx}

\usepackage{adjustbox} 
\usepackage{enumitem}
\usepackage{xcolor}
\usepackage{tabularx}
\usepackage{amssymb}
\usepackage{titlesec}
\usepackage{multirow}
\usepackage{hhline}
\usepackage{natbib}
\usepackage{booktabs}
\usepackage{subcaption}
\usepackage{amsmath}
\usepackage{algorithm}
\usepackage{algorithmic}

\definecolor{lightgreen}{RGB}{180, 230, 180} 

\defcitealias{mathew2021hatexplain}{HateXplain}
\defcitealias{elsherief2021latent}{Latent Hate}

\title{SMARTER: A Data‑efficient Framework to Improve Toxicity Detection with Explanation via Self-augmenting Large Language Models}

\author{Huy Nghiem, Advik Sachdeva, Hal Daumé III  \\
  University of Maryland \\
  \texttt{\{nghiemh, asachde1, hal3\}@umd.edu} 
  }

\setlist{nolistsep,noitemsep,label=$\diamond$}

\begin{document}
\maketitle
\begin{abstract}

\textcolor{red}{\textbf{WARNING:} \textit{This paper contains examples of offensive materials.}} To address toxic content on social media, we introduce \textit{SMARTER}, a data-efficient 2-stage framework for explainable content moderation using Large Language Models (LLMs). In Stage 1, we leverage LLMs' own outputs to generate synthetic explanations for correct and incorrect labels, enabling preference optimization with minimal supervision. In Stage 2, we refine explanation quality through cross-model training, allowing weaker models to align with stronger ones. Experiments on 3 classification tasks (\textit{HateXplain, Latent Hate, Implicit Hate}) show SMARTER achieves up to 13\% macro-F1 improvement over few-shot baselines using only 6-57\% of training data. Our framework offers a scalable strategy for low-data settings by harnessing LLMs' self-improvement for explainable moderation.

\end{abstract}

\section{Introduction}
In recent years, social media platforms have enabled millions of users to virtually interact  at a global scale. While a vital channel to disseminate information and knowledge, these platforms also facilitate the spread of harmful materials \cite{hwang2015social, cinelli2021echo, nguyen2024decade}. Arguably the most concerning among them is toxic content, which could induce serious negative psychological impact on the audience \cite{castano2021internet, windisch2022online, tontodimamma2021thirty}. As a result, content moderation is crucial to detect and mitigate its harm. 

Toxic content encompasses a wide spectrum of terminologies whose definitions vary by platform: hate speech, cyberbully, sexist, racist etc. \cite{gelber2021differentiating, anjum2024hate, wang2025sahsd}. Typically, human moderators must manually adjudicate the target content, an inefficient  process taxing to the moderators themselves \cite{baker2020covid19,  spence2023psychological}. Recently, social media platforms can significantly reduce this overhead by integrating  machine learning models to automate the detection pipeline \cite{d2020bert, sumanth2022toxic}. 

While capable of achieving impressive classification performance, these increasingly sophisticated models require ample resources to train, especially when the target label space covers nuanced concepts \cite{fortuna2021well, sap2021annotators, zhao2021comparative}. Developing a robust dataset for training typically demands significant effort in curation and human annotation, often involving iterative updates to adapt the model to evolving linguistic trends \cite{toraman2022large, bespalov2023towards, rad2025refining}. In addition, this pipeline is limited to decision-only pipeline. Failure to provide users with human-understandable signals renders the adjudication process opaque, inviting further issues on transparency and trustworthiness. 

In this paper, we address the pressing need for intuitive explanations as a cornerstone of content moderation by leveraging Large Language Models (LLMs), which have shown remarkable capabilities in reasoning-related tasks \cite{wei2021finetuned, yao2024tree}. Specifically, our contributions are:

\begin{itemize}
    \item We introduce \textbf{SMARTER}\footnote{Our repository can be found at \url{https://github.com/hnghiem-nlp/hate_dpo_public}} (\underline{S}elf-aug\underline{M}ent\underline{A}tion \underline{R}egimen \underline{T}owards \underline{E}fficient Content Mode\underline{R}ation), a 2-stage framework  where (1) each LLM self-augments on few-shot data using preference optimization with synthetic explanations, (2) cross-model refinement to further improvement by training one LLM with responses generated by another on  held-out sets.
    \item Experiments conducted on 3 classification tasks (\textit{HateXplain, Latent Hate, Implicit Hate}) with Llama-3.1-8B-Instruct and COT-T5-XL demonstrate up to 13.5\% increase in macro-F1 scores in few-shot settings with high-quality explanations, outperforming  a range of commercial models.
\end{itemize}

While commercial APIs may incur high costs and limited transparency, SMARTER delivers competitive performance and explainability with 6-57\% of training data, ensuring deployment control.

\section{Related Work}
\paragraph{Robust Toxicity Detection}
Existing research \cite{caselli2021hatebert, sarkar2021fbert, toraman2022large} has proposed strategies to enhance cross-domain robustness in toxicity detection, such as combining multiple datasets to improve generalization \cite{antypas2023robust} and incorporating label definitions to support few-shot adaptation \cite{nghiem2024define}. Building on these advances, recent studies leverage LLMs for toxicity detection, demonstrating strong zero- and few-shot performance \cite{masud2024hate, jahan2024comprehensive, roy2023probing, kumarage2024harnessing}. LLMs can also generate textual explanations, enhancing interpretability for moderators and users \cite{di2024explanation, yang2023hare, calabrese2024explainability, almohaimeed2025towards}. We extend this line of work by training open-source LLMs to jointly perform classification and explanation to promote transparency.

\paragraph{Aligning LLMs to Human Preferences} In addition to extensive pretraining, modern LLMs also undergo post-training that both unlocks additional capabilities \cite{longpre2023flan} and aligns the outputs towards human preferences \cite{ouyang2022training}. Recent Reinforcement Learning (RL)-based techniques \cite{shao2024deepseekmath, zweiger2025self, nghiem2025balancing} enable promising LLM self-alignment at expensive computational cost. Another body of work showcases LLMs to self-improve via their own synthetic data \cite{tang2023does, li2023synthetic}, such as Self-Refine \cite{madaan2023self} and Self-Instruct \cite{wang2022self}. Building on these ideas, our framework presents a simple yet principled approach  that allows LLMs to iteratively improve with minimal training cost while amenable to human supervision. 
\section{Overview of Framework} 
As shown in Algorithm \autoref{algo:short}, our framework consists of two stages. First, we focus on optimizing the LLMs' classification performance through alignment tuning using self-augmenting synthetic data generation. In the second stage, we refine the quality of the generated explanations by cross-model refinement, leveraging the synergy between different LLMs. Section \ref{sec:improv_cls} and \ref{sec:stage1} present the experimental setup and findings for Stage 1. Section \ref{sec:improv_qual} focuses on  cross-model improvement in Stage 2. Section \ref{sec:nli_eval} analyzes the consistency of explanations with respect to the definitions and labels. 

\begin{figure*}[t]
    \centering
    \includegraphics[width=0.9\linewidth]{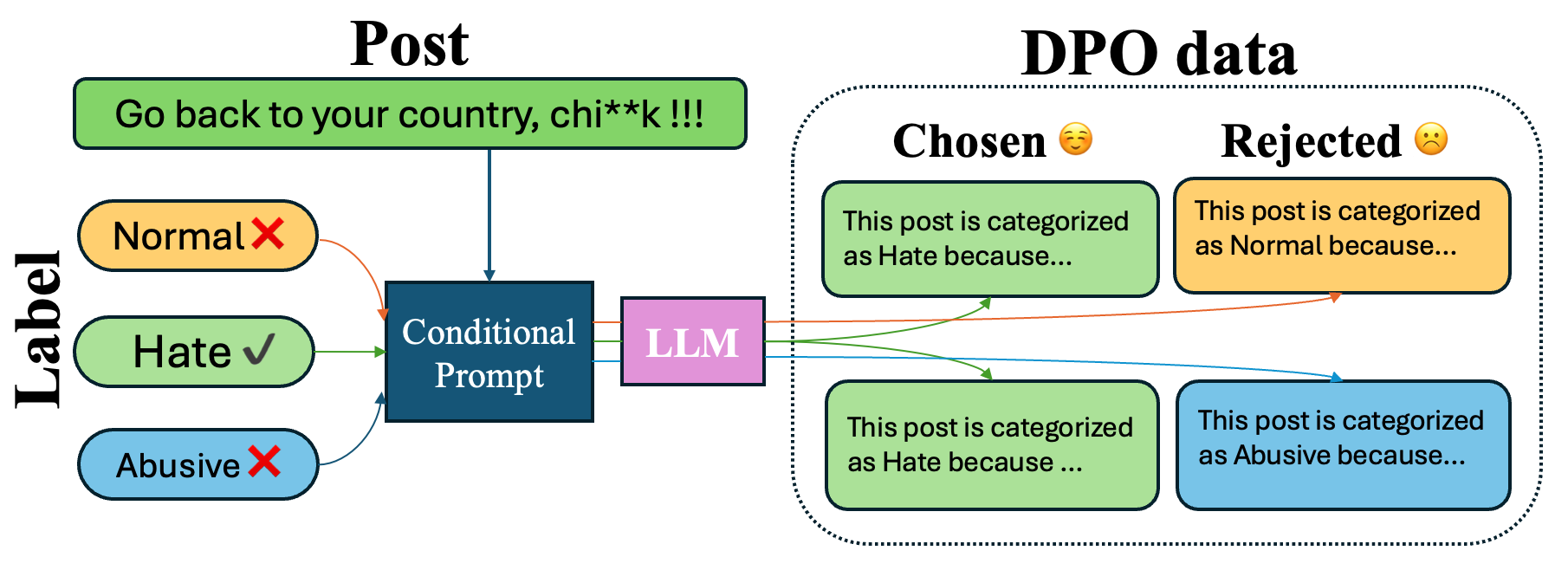}
    \caption{Self-augmenting pipeline: for each post, explanations are conditionally generated for the gold label and all incorrect labels using prompt template in \autoref{fig:cls_prompt}. For DPO, data consists of matching the explanation of the correct label with another incorrect label's. }
    \label{fig:aug_data}
\end{figure*}

\begin{figure}[h]
\centering
\colorbox{blue!10}{
  \parbox{0.95\linewidth}{
    \textbf{Input:} It means people aren't buying the lie! diversity k**s now prove me wrong!

    \textbf{Explanation:} The post uses coded or indirect language to disparage a person or group based on their race. Overall, the use of the phrase "diversity k**s" and the challenge to "prove me wrong" indicate a negative view of diversity, suggesting a belief that diversity is harmful. This coded language indirectly disparages people of color, making it an example of implicit hate. 
    
    \textbf{Label}: Implicit Hate
  }
}
\caption{Example of explanation and classification label produced by \textit{Llama} at $K=128$ shot on the dataset \textit{Latent Hate}. For definition of categories, see \autoref{tab:hate_definitions}.  }
\label{fig:exp_example}
\end{figure}

\begin{algorithm}[t]
\small
\caption{High-Level Overview of SMARTER}
\label{algo:short}
\begin{algorithmic}[1]

\STATE \textbf{Input:} LLMs $\mathcal{L}$,  performance metric $\mathcal{M}$ \\
Datasets $\mathcal{D}_{\text{train}} \text{ (with explanation)}, \mathcal{D}_{\text{val}}, \mathcal{D}_{\text{test}}$, 
\STATE \textbf{Optional:} Initial SFT pretraining on relevant data

\vspace{0.5em}
\STATE \textbf{Stage 1: Individual LLM Self-Augmentation}
\FOR{$K \in \{K_{\text{start}},...,K_{\text{end}}$ \}} 
    \STATE Sample $K$-shot subset $\mathcal{D}^{(K)}$
    \FORALL{$\text{LLM}_i \in \mathcal{L}$}
        \STATE SFT on $\mathcal{D}^{(K)}$
        \STATE Collect \textit{chosen} responses for gold labels
        \STATE Collect \textit{rejected} responses for incorrect labels
        \STATE Align via preference optimization using synthetic preference data; evaluate on $\mathcal{D}_{\text{val}}$
    \ENDFOR
\ENDFOR

\vspace{0.5em}
\STATE \textbf{Stage 2: Cross-Model Refinement}
\FOR { $\text{LLM}_a \ne \text{LLM}_b \in \mathcal{L}$}
    \STATE Select $K_{check} \in  \{K_{\text{start}},...,K_{\text{end}}\} \rightarrow \mathcal{D}^{(K_{check})} $
    \STATE SFT $\text{LLM}_a$ on $\mathcal{D}^{(K_{check})} $
    \STATE Self-augment $\text{LLM}_a$ with responses from $\text{LLM}_b$ on $\mathcal{D}^{(K_{check}')} \ne \mathcal{D}^{(K_{check})} $
\ENDFOR

\RETURN Model with best performance $\mathcal{M}$ on $\mathcal{D}_{val/test}$
\end{algorithmic}
\end{algorithm}

\section{Preliminary Setup}
\label{sec:improv_cls}
In this section, we provide details about the empirical setup on data, model, and a description of classification technique via prompting. 

\subsection{Data}
We select 2 datasets that cover distinct yet still relevant concepts to toxic content moderation and derive from them 3 classification tasks. 
\begin{itemize}
    \item \textbf{HateXplain}, introduced by \citet{mathew2021hatexplain}. Consisting of posts collected from 2 social media sites Gab and Twitter (now X), this dataset contains 20,148 samples annotated by crowdworkers of the Amazon Mechanical Turks 
    (AMT) platform on 3 labels \textit{Hate, Offensive} and \textit{Normal}.
    \item \textbf{Latent Hate}, curated by  \citet{elsherief2021latent},  focuses on covert hate speech collected from Twitter. With 19,112 samples also annotated by AMT workers, this dataset's primary label space contains 3 high-level categories: \textit{Not Hate, Implicit Hate, Explicit Hate}. The authors also assigned the subset of 4,153 tweets of the \textit{Implicit Hate} category a label among 6  secondary fine-grained categories: \textit{White Grievance, Incitement to Violence, Inferiority Language, Irony, Stereotypes and Misinformation, Threatening and Intimidation}. 
\end{itemize}
Label \textit{definitions} are included, enabling context integration into explanations following \citet{yang2023hare} and \citet{nghiem-daume-iii-2024-hatecot}. We denote these three tasks as \textit{HateXplain}, \textit{Latent Hate}, and \textit{Implicit Hate}. \autoref{tab:test_data} summarizes data splits; preprocessing details are in Appendix~\ref{apx:data}. \autoref{fig:exp_example} provides an example of the explanation produced by our model on \textit{Latent Hate}.

\subsection{Models}
Extending our focus on accessibility, we employ 2 open-source LLMs that have exhibited strong performance in similar classification tasks: 
\begin{itemize}
    \item \textbf{COT-T5-XL}: A variant of the base encoder-decoder Flan-T5-XL \cite{longpre2023flan} architecture, this model consists of 3 billion parameters and is further trained on the COT corpus, a collection of 1.8 million samples with chain-of-thought style explanations, which enhances its capabilities on a variety of reasoning tasks \cite{kim2023cot}.
    \item \textbf{Llama-3.1-8B-Instruct}: A decoder-only LLM released by Meta \cite{dubey2024llama}. Consisting of 8 billion parameters, this model is further instruction-tuned to aligned with human preferences in addition to being trained on extensive linguistic corpora.  
\end{itemize}
For brevity, we refer to the models as \textit{T5} and \textit{Llama} respectively for the remainder of this paper.

\subsection{Classification via Prompting} We apply the Chain-of-Thought (COT) paradigm \cite{wei2022chain} in our classification tasks to obtain the explanation from the LLMs necessary for content moderation. Inspired by \citet{nghiem-daume-iii-2024-hatecot} and \citet{yang2023hare}, we use the prompt in \autoref{fig:cls_prompt}, which incorporates the target post, the set of labels corresponding to the task and their associated definitions as context for the LLM to provide its explanation and predicted category. 

\subsection{Auxiliary Pretraining} 
\citet{nghiem-daume-iii-2024-hatecot} found that supervised finetuning (SFT) off-the-shelf LLMs on \textit{HateCOT} -- a general compilation of toxic posts, labels, definitions and synthetic explanations -- can boost their performance in downstream hate speech detection. To maximize expected gains with minimal in-domain data, we adopt this strategy and perform SFT on the 2 chosen LLMs using \textit{HateCOT} to prime them for our classification tasks. Finetuning is performed by training LoRA adapters \cite{hulora}, a parameter-efficient approach implemented by the HuggingFace library  (see \autoref{apx:implement}) as the basis for downstream experiments. 

\section{Stage 1: Individual Model Training}
\label{sec:stage1}
We demonstrate SMARTER's data efficiency by training models at extreme low-shot settings and evaluating against off-the-shelf, commercial, and full-training baselines.

\subsection{K-shot In-domain Finetuning}
\label{sec:kshot}
For each task, we select uniformly at random K-shot without replacement $ \in \{16, 32, 64, 128, 256\}$ samples from the training set . We utilize the seed explanations by \citet{nghiem-daume-iii-2024-hatecot} to further SFT the existing \textit{HateCOT}-pretrained  \textit{T5} and \textit{Llama} models  similar to the auxiliary pretraining phase above.

\paragraph{Self-augmenting via Alignment Tuning} \label{sec:self-aug} We optimize the LLMs' performance while not having access to more training data by exploiting the models' sycophantic tendency. More concretely, we augment the training sample size by prompting the LLMs to generate explanations \textit{
conditioned on the incorrect labels} (as illustrated in \autoref{fig:aug_data}). This process supplies us with synthetic post-explanation pairs that are by design \textit{dispreferred} compared to the true \textit{preferred} sample pairs, enabling the use of the following alignment techniques to further finetune the models. 

\begin{itemize}
    \item \textbf{Direct Preference Optimization (DPO)}: An offline RL-based technique that optimizes the policy via an implicit reward model using preference data \cite{rafailov2024direct}. DPO requires human preference data where a pair of chosen vs. rejected responses is presented for each prompt. 
    
    \item \textbf{Kahneman-Tversky Optimization (KTO)}: An alternative to DPO, KTO leverages prospect theory to construct human-aware losses (HALOs) to optimize the policy \cite{ethayarajh2024kto}. KTO only requires listwise human preference data: a data point consists of a prompt, a response and a binary flag that indicates the acceptability of the response.
\end{itemize}

Appendix \ref{apx:alignment} offers more technical details on these techniques. Our self-augmenting methodology naturally procures training datasets in the required format for KTO by simply designating the explanation conditioned on the gold label as positive, and otherwise negative. For DPO, we match the explanations conditioned on the correct-incorrect pair sequentially to satisfy the pairwise chosen-rejected format. As an example, 100 posts of \textit{HateXplain} (whose label space is of size 3), should produce $100 * 3=300$ training samples for KTO, and $100 * (3 - 1) = 200$ for DPO.

\subsubsection{Experimental Pipeline}
\label{cls_pipeline}

The experiments follow a two-stage process. We first finetune \textit{T5} and \textit{Llama} at $K \in \{16,32,64,128\}$ followed by DPO self-augmentation. At $K=256$, we compare KTO and DPO alignment.\footnote{KTO trainer for \textit{T5} is unavailable, so we evaluate KTO only for \textit{Llama}.} For comparison, we finetune both LLMs on full training data for classification without explanations (\textit{Full} models), and similarly train ModernBERT (large)~\cite{warner2024smarter} as an additional baseline.

In addition, we examine 2 partial sampling strategies to circumvent the proportionate growth of self-augmenting synthetic data to label space when K is large . In the first, we construct augmented data by selecting 128 and 192 shots of samples uniformly at random from the Baseline's 256-shot pool. In the second, instead of $K$-shot sampling, we select $K' * |S| * (|S| -1)$ samples from all possible post-label pairs of the 256-shot Baseline pool (incorrect labels included) to create augmented data, where $K' \in \{128, 192\}$ and $|S|$ is the label space's size. While both approaches should yield identical size of training data for the same $K'$, the first method, denoted \textit{DPO-K}, preserves post-label parity whereas the second method, denoted \textit{DPO-N}, yields higher diversity of post content. 

Experiments are implemented with the HuggingFace library. We sample $K=50$ shot on the held out Validation extraction of each dataset to tune hyperparamters, and evaluate on samples from the Test set as shown in \autoref{tab:test_data}. Technical details are included in Appendix \ref{apx:implement}.

\subsection{Classification Results}
\subsubsection{Results for $K <= 128$}\label{sec:dpo128} 
\begin{figure*}[!ht]
    \centering
    \captionsetup{aboveskip=0pt, belowskip=0pt} 
    \setlength{\belowcaptionskip}{-5mm} 

    \begin{subfigure}{\linewidth}
        \centering
        \includegraphics[width=0.48\linewidth]{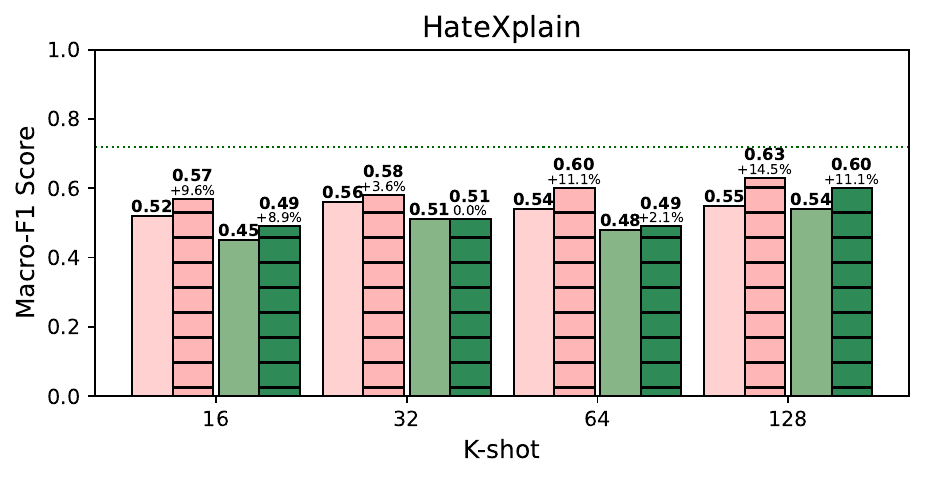}
        \includegraphics[width=0.48\linewidth]{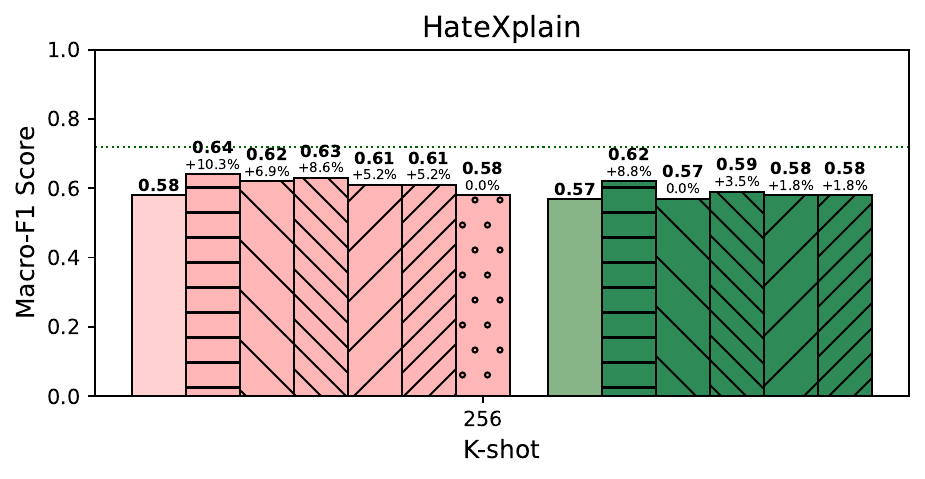}
        \label{fig:cls_1}
    \end{subfigure}
    \vspace{-5mm} 

    \begin{subfigure}{\linewidth}
        \centering
        \includegraphics[width=0.48\linewidth]{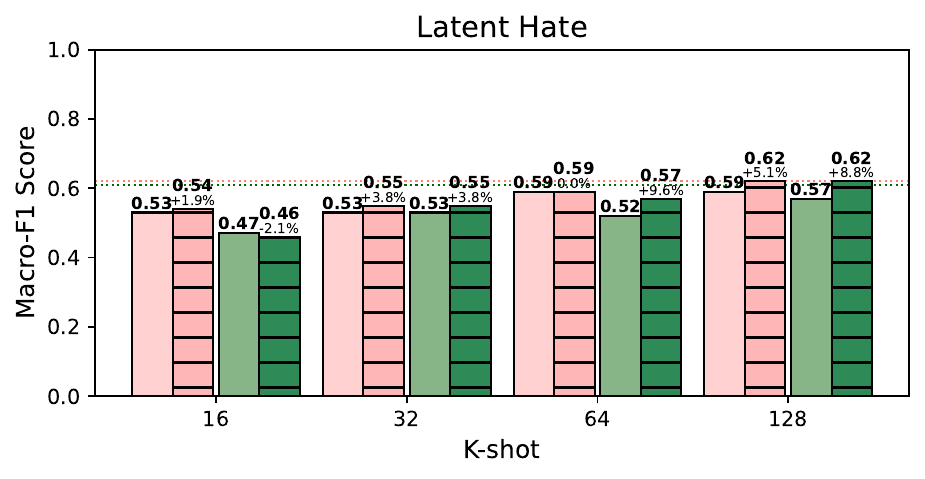}
        \includegraphics[width=0.48\linewidth]{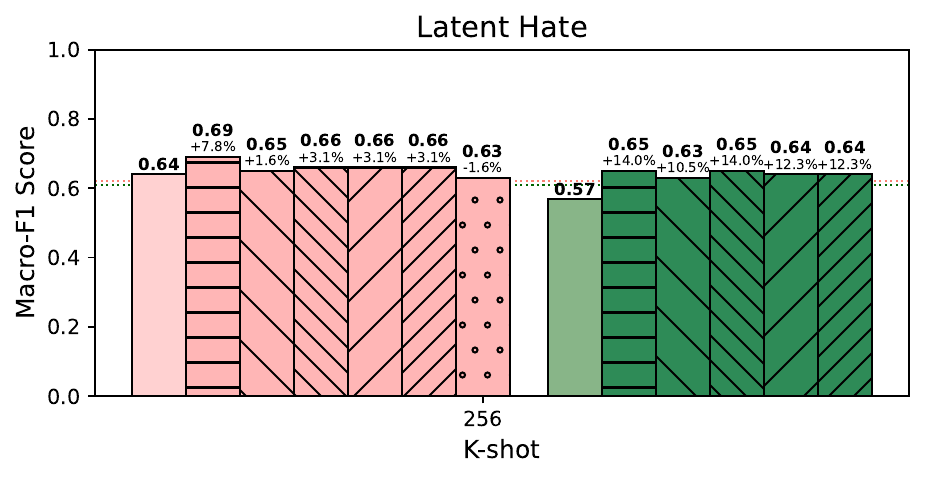}
        \label{fig:cls_2}
    \end{subfigure}
    \vspace{-5mm} 

     \begin{subfigure}{\linewidth}
        \centering
        \includegraphics[width=0.48\linewidth]{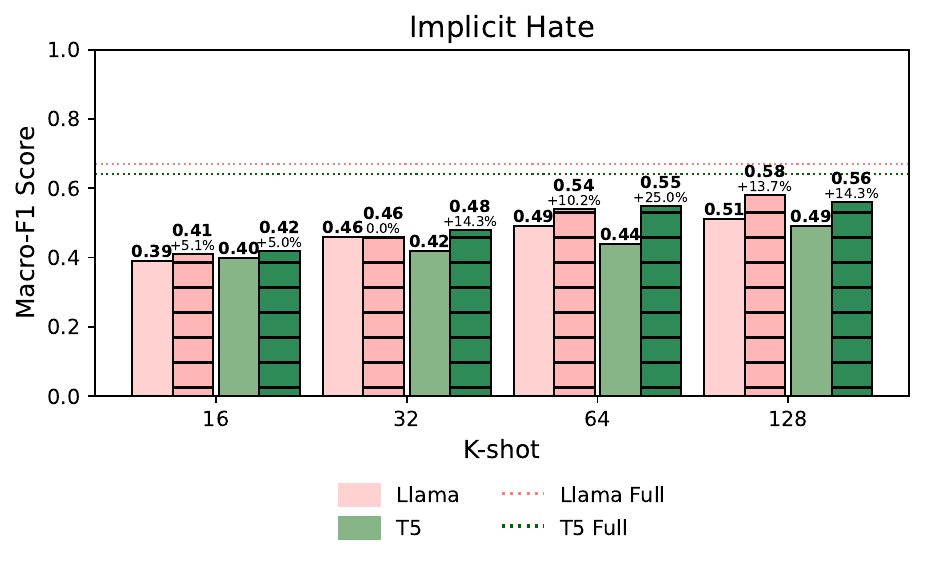}
        \includegraphics[width=0.48 \linewidth]{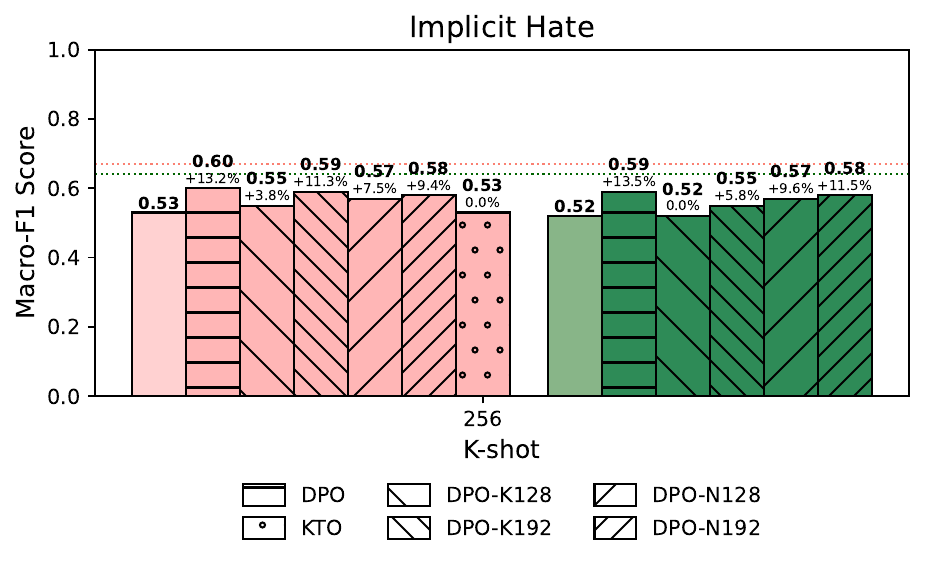}
        \label{fig:cls_3}
    \end{subfigure}
    \vspace{-2mm} 
    \caption{Bar plots for $K$-shot classification experiments on 3 datasets using \textit{Llama} and \textit{T5} models. Macro-F1 scores and percentage change over Baseline are displayed on top. Results for Baselines and DPO-augmented variants for $K \in \{16, 32, 64, 128\}$ are displayed on the left subfigures. Results for $K=256$ of Baseline, KTO, DPO-augmented and its other variants on various sub-sampling strategies (section \ref{cls_pipeline}) are shown on the right. Horizontal lines show the F1 scores for \textit{Full} models that use all training data.}
    \label{fig:cls_result}
    
\end{figure*}

We report macro-F1--unweighted mean of per-class F1 scores--as the evaluation metric \cite{christen2023review}. \autoref{fig:cls_result} displays the bar plots of the LLM's Baseline and DPO-enhanced performance on the test sets.

Self-augmentation with DPO generally boosts classification performance over Baseline counterparts, with the sole reversal for \textit{T5} on \textit{Latent Hate} at $K=16$ (2.1\% reduction), along with static scores for \textit{T5} on \textit{HateXplain} at $K=32$ and \textit{Llama} on \textit{Latent Hate} at $K=64$. These infrequent and negligible exceptions highlight our method's consistent enhancement. For instance,  F1 gains reach 14.5\% for \textit{Llama} (\textit{HateXplain, $K=128$}) and 25\% for \textit{T5} (\textit{Implicit Hate}, $K=64$).

Notably, DPO self-augmentation improves performance even at $K \le 64$, showcasing our method's advantage in low-resource settings. \textit{Llama} Baseline models appear to consistently outperform \textit{T5} across all tasks. However, with DPO self-augmentation, \textit{T5 } is able to  narrow the margin on \textit{Latent Hate} and \textit{Implicit Hate}. In contrast, \textit{Llama} still dominates on \textit{HateXplain}. These discrepancies suggest an interaction between dataset and model choice, and practitioners may benefit from contrasting different options with our self-augmenting techniques for best results.

\subsubsection{Results for $K=256$}

In \autoref{fig:cls_result}, we observe the macro-F1 scores for models $K=256$ on the subfigures on the right. DPO-augmentation continues to enhance Baseline models which already benefits from more training data relative to lower $K$ values, with improvement in scores reaching up to 13.2\% for \textit{Llama} and 14\% for \textit{T5}. In contrast, KTO-augmentation appears ineffectual on \textit{HateXplain} and even hinders performance on \textit{Latent Hate} and \textit{Implicit Hate}, showing notable reduction from Baseline scores, respectively. With this, we hypothesize that DPO's contrastive nature reinforces signals that allow the LLM to learn from its own mistakes, thereby sharpening its reasoning ability for this type of discriminatory task. On the other hand, KTO's reliance on implicit signals from listwise data is not sufficient to improve discernment of classes when the semantic distinction between labels are nuanced for \textit{Latent Hate} and \textit{Implicit Hate}.

Results also show that DPO-augmented sub-sampling strategies still demonstrate generally better F1 scores than Baselines, further reinforcing the benefit of DPO-augmentation. There appears no clear winner between \textit{DPO-K} and \textit{DPO-N} among all 3 datasets. Augmenting with larger $K'$ values (approaching 256) consistently yields scores closer to the maximal F1, indicating its role as a hyperparameter that practitioners need to decide for the trade-off between computational cost and benefit.

\begin{table*}[!t]
    \centering
    \footnotesize
    \begin{tabular}{l c c c c c c c c c}
    \toprule[1pt]
        & \multicolumn{3}{c}{\textbf{HateXplain}} & \multicolumn{3}{c}{\textbf{Latent Hate}} & \multicolumn{3}{c}{\textbf{Implicit Hate}} \\
        \cmidrule(lr){2-4} \cmidrule(lr){5-7} \cmidrule(lr){8-10}
        \textbf{Model} & F1 & F1\% & Data\% & F1 & F1\% & Data\% & F1 & F1\% & Data\% \\
        \hline
        \multicolumn{10}{l}{\textit{SMARTER (Ours)}} \\
        \quad Llama\_DPO-256  & 0.64 & 89\% & 6\% & \textbf{0.69} & 100\% & 7\% & 0.60 & 90\% & 57\% \\
        \quad T5\_DPO-256     & 0.62 & 86\% & 6\% & 0.65 & 94\% & 7\% & 0.59 & 88\% & 57\% \\
        \hline
        \multicolumn{10}{l}{\textit{Full Training Baselines}} \\
        \quad Llama\_Full     & \textbf{0.72} & 100\% & 100\% & 0.62 & 90\% & 100\% & \textbf{0.67} & 100\% & 100\% \\
        \quad T5\_Full        & \textbf{0.72} & 100\% & 100\% & 0.61 & 88\% & 100\% & 0.64 & 96\% & 100\% \\
        \quad ModernBERT      & 0.70 & 97\% & 100\% & 0.61 & 88\% & 100\% & 0.64 & 96\% & 100\% \\
        \hline
        \multicolumn{10}{l}{\textit{Commercial: Zero-shot}} \\
        \quad GPT-4o-mini     & 0.50 & 69\% & -- & 0.54 & 78\% & -- & 0.42 & 63\% & -- \\
        \quad GPT-4.1         & 0.55 & 76\% & -- & 0.60 & 87\% & -- & 0.57 & 85\% & -- \\
        \quad GPT-5-chat      & 0.56 & 78\% & -- & 0.51 & 74\% & -- & 0.58 & 87\% & -- \\
        \quad Qwen-32B        & 0.54 & 75\% & -- & 0.47 & 68\% & -- & 0.49 & 73\% & -- \\
        \hline
        \multicolumn{10}{l}{\textit{Commercial: 16-shot ICL}} \\
        \quad GPT-4o-mini     & 0.29$\pm$0.10 & 40\% & -- & 0.25$\pm$0.09 & 36\% & -- & 0.15$\pm$0.01 & 22\% & -- \\
        \quad GPT-4.1         & 0.52$\pm$0.07 & 72\% & -- & 0.63$\pm$0.05 & 91\% & -- & 0.38$\pm$0.11 & 57\% & -- \\
        \quad GPT-5-chat      & 0.62$\pm$0.01 & 86\% & -- & 0.60$\pm$0.06 & 87\% & -- & 0.40$\pm$0.11 & 60\% & -- \\
        \quad Qwen-32B        & 0.55$\pm$0.04 & 76\% & -- & 0.57$\pm$0.03 & 83\% & -- & 0.48$\pm$0.04 & 72\% & -- \\
        \bottomrule[1pt]
    \end{tabular}
    \caption{Comparison of performance and data usage across training paradigms. SMARTER models use DPO with $K{=}256$ shots; commercial models evaluated in zero-shot and 16-shot in-context learning (ICL, mean$\pm$std over 3 seeds). We report macro-F1 (\textbf{F1}), relative score (\textbf{F1\%}), and training data fraction (\textbf{Data\%}). Models with the \textsc{DPO} suffix are DPO-augmented, while \textit{Full} uses the entire training set. Within each dataset, best \textbf{F1} is bolded and \textbf{F1\%} is normalized to that best model (100\%). \textbf{Data\%} is measured relative to the corresponding \textit{Full} baseline (100\%). ModernBERT is shown for reference.}
    \label{tab:compare}
\end{table*}

\paragraph{Ablation and Baseline Comparisons.} To isolate SMARTER's contribution, \autoref{tab:ots_f1} shows off-the-shelf models without HateCOT pretraining achieve 0.52 F1 on \textit{HateXplain}. Adding HateCOT (Baseline $K=256$) improves to 0.58 (+0.06), while SMARTER's DPO self-augmentation reaches 0.64 (+0.06 additional), contributing performance gain.

We evaluate commercial models (GPT-4o-mini, GPT-4.1, GPT-5-chat, Qwen-32B) under zero-shot and 16-shot ICL (3 seeds; \autoref{tab:icl_full}). In \autoref{tab:compare}, our DPO-augmented models at $K=256$ consistently outperform both paradigms. Remarkably, 16-shot ICL often degrades performance: GPT-4o-mini drops from 0.50$\rightarrow$0.29 on \textit{HateXplain} with high variance ($\pm$0.10), while even GPT-5-chat (0.62 mean) trails our Llama\_DPO-256 (0.64). Following \textit{ModernBERT}'s setup, we also train base models without explanations. At $K=256$, our DPO variants retain 86\%+ of \textit{Full} performance on \textit{HateXplain} and surpass it on \textit{Latent Hate}.
 
\section{Stage 2: Cross-model Refinement}
\label{sec:improv_qual}
Having shown individual models self-improve efficiently, we explore whether cross-model training can further enhance performance by transferring explanation quality between LLMs. We first conduct human evaluation of both models' explanations, then train each model on the other's outputs.

\begin{table}[t]
\footnotesize
\centering
\begin{tabular}{lrrr}
\toprule
\textbf{Label} & \textbf{T5} & \textbf{Llama} & \textbf{Row Total} \\
\midrule
Normal             & 25 (25.5\%)      & 73 (74.5\%)        & 98           \\
Offensive          & 63 (49.6\%)      & 64 (50.4\%)        & 127          \\
Hate    & 54 (46.2\%)      & 63 (53.8\%)        & 117          \\
\bottomrule
\end{tabular}
\caption{Annotator preferences for explanations on 342 \textit{HateXplain} samples (majority vote).}
\label{tab:mturk_result}
\end{table}

\subsection{Human Evaluation}
\label{sec:human_val}
Due to resource constraints, we only perform this evaluation pipeline for \textit{HateXplain}. We opt to collect outputs from the SFT+DPO models at \textbf{$K=128$} to preserve the rest of the training data for further refinement experiments. To make a fair comparison on the outputs to agree on the predicted label, we use the prompt in \autoref{fig:cond_prompt} to obtain explanations conditioned on the correct gold label. This technique allows us to obtain 342 samples with consistent labels, or 89\% of the possible $128*3=384$ possible samples. The other 42 samples contain discrepant predicted labels, where the LLMs revert to behaviors ingrained by previous training.
 
 To simulate real-life deployment, we solicit 14 crowdworkers from diverse demographic backgrounds on the platform Amazon Mechanical Turks (AMT) to annotate their preference on the explanations offered by the LLMs in this study . Annotators are encouraged to evaluate the pairs of explanation for each post based on the criteria of \textit{Clarity, Reasoning} and \textit{Alignment} as illustrated with the template in \autoref{fig:mturk}. To avoid position bias, we present the pairs of explanations in randomized orders. All annotators are compensated fairly.

\paragraph{Annotation Results} \autoref{tab:mturk_result} shows the annotation results, aggregated per sample by majority voting. Across the 3 labels in \textit{HateXplain},  annotators show no significant preference for explanations by either model on \textit{Offensive} and \textit{Hate}. On the other hand, \textit{Llama}'s explanations for \textit{Normal} posts are overwhelmingly preferred. 

\subsection{Cross-model Refinement}
Motivated by \textit{Llama}'s superior human-rated explanations, we investigate whether models can benefit from each other's outputs through cross-training. Below we describe our cross-model methodology, evaluate its impact on classification accuracy, and analyze how explanation styles evolve.

\subsubsection{Cross-model Training Methodology}
While \textit{Llama} produces human-preferred explanations and is the natural deployment choice, we investigate whether models can adopt each other's explanation styles without degrading classification performance. Using the prompting technique described in Section \ref{sec:human_val}, we use the SFT+DPO $K=128$ variants of both \textit{Llama} and \textit{T5} LLMs to generate explanations conditioned on the gold labels of the held-out 128-shot portion of the training data (unseen by either model). We then perform cross-model training: \textit{T5} is finetuned on \textit{Llama}'s outputs and vice versa, applying both SFT and DPO self-augmentation as in Section~\ref{sec:kshot}.

\subsubsection{Cross-model Classification Results}
\begin{figure}[t]
    \centering
    \includegraphics[width=0.95\linewidth]{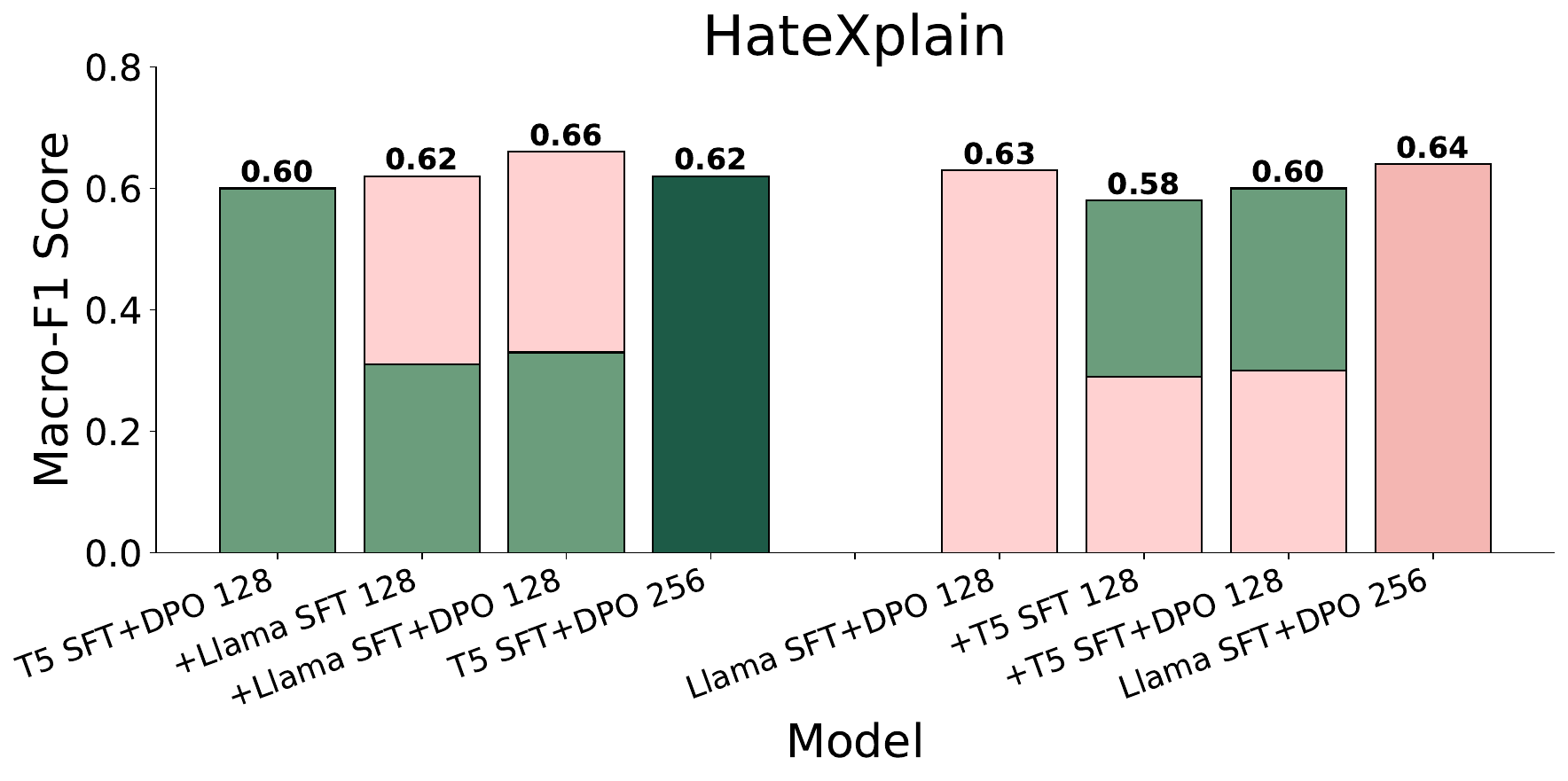}
    \includegraphics[width=0.95\linewidth]{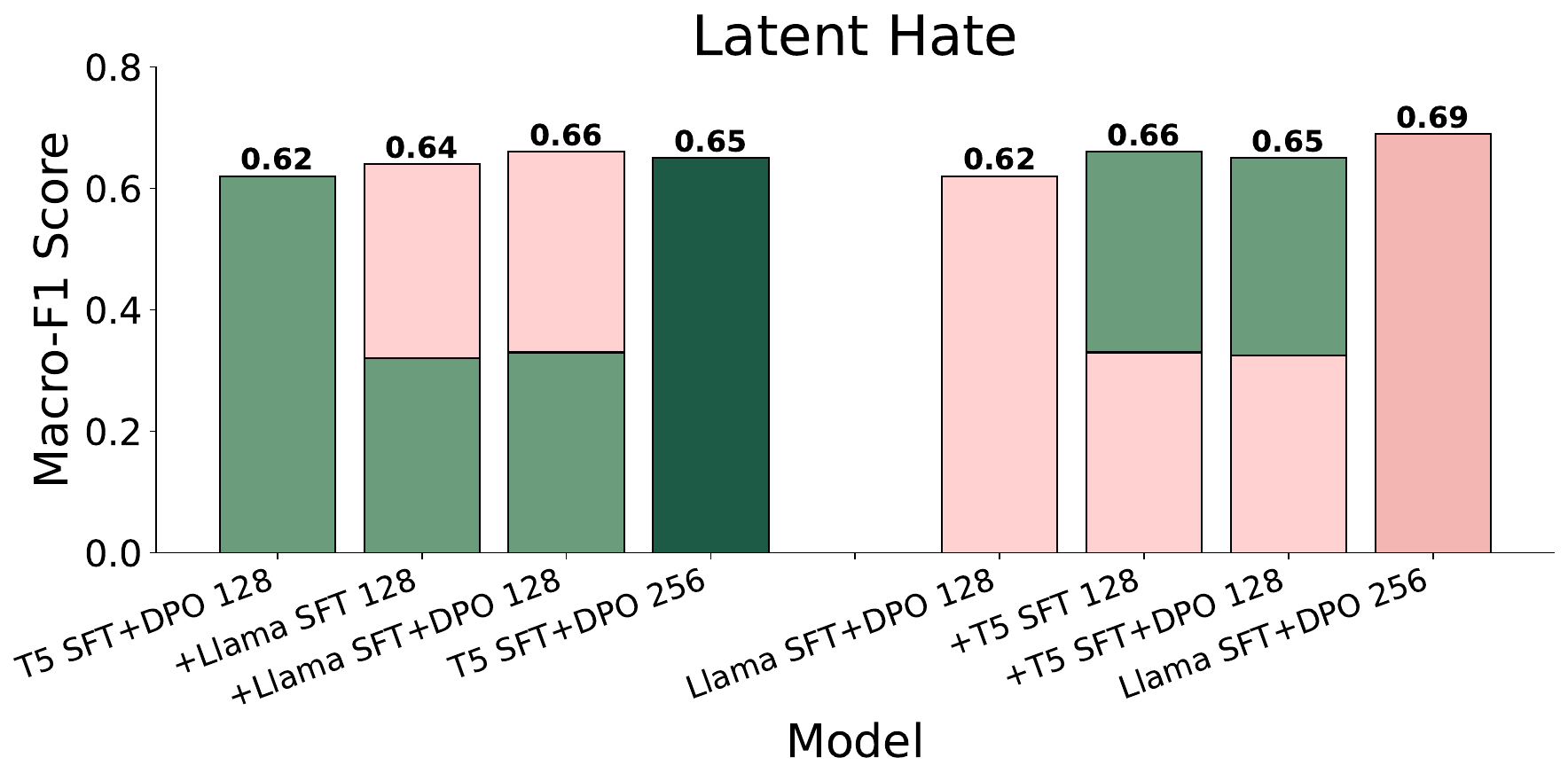}
    \includegraphics[width=0.95\linewidth]{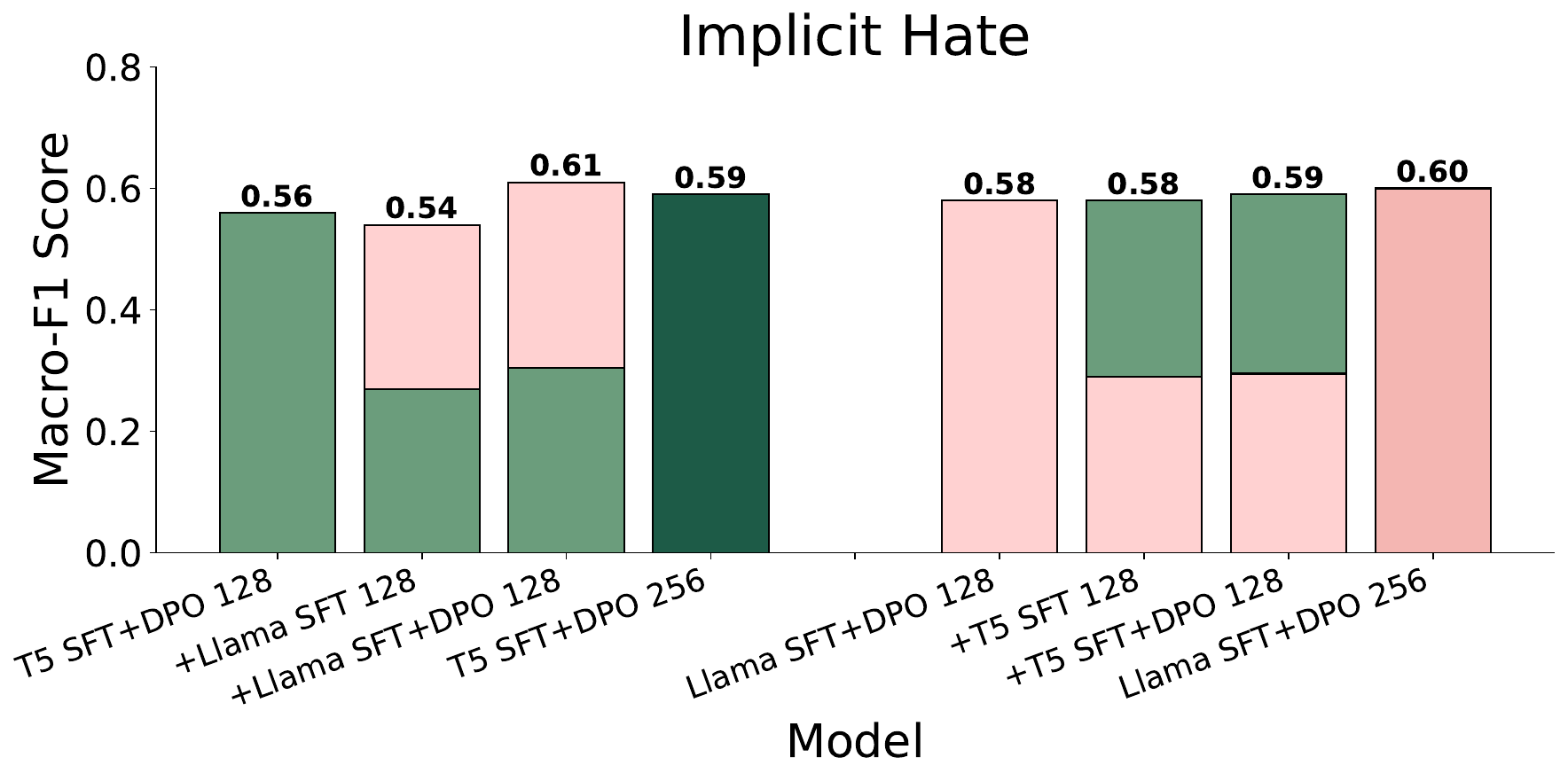}
    \caption{Macro-F1 scores on test portion of the 3 test sets for \textit{T5} and \textit{Llama} cross-model refinement regimen. In each figure: the first and last bars are scores reported with only \textit{T5} model at $K=128$ and $K=256$ (directly from \autoref{fig:cls_result}); the middle bars are results after further finetuned using data from the complementary $K=128$ shots of counterpart model. Split-color bars: bottom color indicates the original model; top color indicates the counterpart model for additional cross-training.  }
    \label{fig:x_refine}
\end{figure}

\autoref{fig:x_refine} shows the macro-F1 scores of both the original single-model SFT+DPO and cross-model finetuned variants on the test datasets. 

\paragraph{Cross-model training enhances T5 while impairing Llama.} Applying SFT training using explanations from \textit{Llama} improves performance over the baseline \textit{T5} SFT+DPO on \textit{HateXplain} (F1 scores 0.62 from 0.60) and \textit{Latent Hate} (0.64 from 0.62), with a slight reduction on \textit{Implicit Hate} (0.54 from 0.56). On the other hand, applying SFT+DPO augmentation using \textit{Llama}'s explanations from the complementary K=128 shot data allows T5 to  \textbf{exceed} single-model performance on the full K=256 set across the board. The new maximum macro-F1 scores for \textit{T5} are 0.66 on \textit{HateXplain} (up from 0.62 of single-model), 0.66 on \textit{Latent Hate} (up from 0.65), and 0.61 on \textit{Implicit Hate} (up from 0.59). Notably, these scores are superior to even \textit{Llama}'s single-model performance at $K=256$ on \textit{HateXplain} and \textit{Implicit Hate}.

In contrast, SFT cross-training with \textit{T5}'s output reduces \textit{Llama}'s F1 scores to 0.58 from 0.63 on \textit{HateXplain} while improving to 0.66 from 0.62 on \textit{Latent Hate} and plateauing at 0.58 on \textit{Implicit Hate}. Additional SFT+DPO cross-model augmentation slightly boosts performance on \textit{HateXplain} and \textit{Implicit Hate} while slightly impairing on \textit{Latent Hate}. None of the cross-model \textit{Llama} variants matches the single-models' F1 scores at $K=256$.

\subsubsection{Cross-model Style Analysis}

\begin{table*}[t]
\footnotesize
\centering
\caption{NLI-based consistency between explanations, predicted labels, and label definitions across datasets. Both DPO and cross-refined (XMOD) models show high entailment rates (>96\%) with marginal percentages of contradictions (Contra.), neutral or undefined (Undef.) edge cases. Categories with all $0$ values are omitted.}
\label{tab:nli}
\setlength{\tabcolsep}{6pt}
\renewcommand{\arraystretch}{1.15}
\begin{tabular}{lll
                *{3}{r}
                *{3}{r}}
\toprule
\multicolumn{3}{c}{} &
\multicolumn{3}{c}{\textbf{Label Consistency (\%)}} &
\multicolumn{3}{c}{\textbf{Definition Consistency (\%)}} \\
\cmidrule(lr){4-6}\cmidrule(lr){7-9}
\textbf{Dataset} & \textbf{Model} & \textbf{Train} &
\textbf{Entail} & \textbf{Contra.} & \textbf{Undef.} &
\textbf{Entail} & \textbf{Contra.} & \textbf{Neutral} \\
\midrule
\multirow{4}{*}{\textit{HateXplain}}
  & \multirow{2}{*}{T5}   & DPO     & 99.2 & 0.8 & 0.0 & 98.2 & 1.5 & 0.3 \\
  &                        & XMOD  & 96.7 & 1.5 & 1.8 & 99.0 & 0.9 & 0.2 \\
  & \multirow{2}{*}{Llama} & DPO     & 97.5 & 1.2 & 1.3 & 97.2 & 2.6 & 0.2 \\
  &                        & XMOD  & 96.4 & 3.6 & 0.0 & 96.7 & 3.2 & 0.1 \\
\midrule
\multirow{4}{*}{\textit{Latent Hate}}
  & \multirow{2}{*}{T5}   & DPO     & 98.8 & 1.2 & 0.0 & 97.6 & 2.4 & 0.0 \\
  &                        & XMOD  & 97.4 & 1.9 & 0.7 & 97.6 & 2.3 & 0.1 \\
  & \multirow{2}{*}{Llama} & DPO     & 97.3 & 2.7 & 0.0 & 97.3 & 2.3 & 0.3 \\
  &                        & XMOD  & 96.8 & 3.2 & 0.0 & 97.5 & 2.5 & 0.0 \\
\midrule
\multirow{4}{*}{\textit{Implicit Hate}}
  & \multirow{2}{*}{T5}   & DPO     & 99.6 & 0.4 & 0.0 & 99.0 & 1.0 & 0.0 \\
  &                        & XMOD  & 98.3 & 1.4 & 0.4 & 97.5 & 2.0 & 0.5 \\
  & \multirow{2}{*}{Llama} & DPO     & 98.1 & 0.4 & 1.5 & 97.5 & 1.8 & 0.8 \\
  &                        & XMOD  & 96.9 & 2.8 & 0.1 & 96.0 & 3.6 & 0.4 \\
\bottomrule
\end{tabular}
\end{table*}
We additionally train a style classifier based on the BERT architecture using explanations generated individually by \textit{T5} and \textit{Llama} of the SFT+DPO variant at $K=128$ \cite{devlin2019bert}. A classification head is added to BERT, which leverages the \textit{[CLS]} token from each explanation's embedding to predict whether it was produced by \textit{T5} or \textit{LLaMA}. Technical details are reported in Appendix \ref{apx:alignment}.

\autoref{fig:bert_datasets} shows individually trained models exhibit their own styles, while cross-model SFT aligns outputs with the alternative's style. Except for \textit{T5} on \textit{HateXplain}, cross-model DPO slightly amplifies the base model's style, suggesting this mechanism induces mild stylistic convergence.

\paragraph{Cross-model explanations may improve weak categories in weaker models.} We compare \textit{Precision}, \textit{Recall}, and \textit{F1-score} of base and cross-trained models at $K=256$ (\autoref{fig:xmodel_anl}) to assess cross-model refinement. Improvements often occur in categories where the paired model excels, indicating transferable reasoning patterns, while declines appear when the counterpart is weak. This asymmetry underscores our recommendation in line~19 of Algorithm~\ref{algo:short} to validate and select the final model using a robust performance metric.

\section{Analysis of Explanations}
\label{sec:nli_eval}
Using Natural Language Inference (NLI), we assess explanation consistency with labels and definitions, then analyze cross-model stylistic changes.

\subsection{Analysis of Consistency}

\paragraph{Explanation-Label Consistency} We prompt GPT-4o-mini with the template shown in \autoref{fig:nli_prompt_1} to judge whether the offered explanations are logically consistent with the predicted label for each sample across 3 test sets via the 4 categories: Entail, Contradict, Neutral and Undefined (noise cases where either the predicted label or the explanation is missing). An author's independent annotation of 150 random samples from this pool under the same instruction achieves nearly unanimous agreement with GPT-4o-mini (Cohen's $\kappa = 0.92$), an observation inline with recent works that validate LLM-as-a-judge approach in NLI applications \cite{negru2025morphnli, gallipoli2025not}.

In \autoref{tab:nli}, Entail values exceeding 96\% indicate that \textit{explanations are largely consistent with their respective labels}. However, cross-model training induces a marginal 2-3\% increase in Contradiction compared to the DPO augmentation.

\paragraph{Explanation-Definition Consistency} Applying the same method to \textit{only the Entail samples} from the previous analysis, using the prompt shown in \autoref{fig:nli_prompt_2} (Cohen’s $\kappa = 0.96$), we again observe consistently high entailment rates and a slight increase in contradiction under cross-model training that mirrors the aforementioned trend. 

\textit{Overall, LLM-generated explanations remain predominantly consistent with both predicted labels and their definitions.} Cross-model refinement can enhance classification but may introduce a slight risk of inconsistency, warranting periodic monitoring. Appendix \ref{apx:nli_edge} offers qualitative analysis and insights for further quality checks for deployment.

\section{Discussion and Conclusion}
Our empirical analysis yields several key recommendations to guide practitioners in deployment.

\paragraph{SMARTER delivers controllable, cost-efficient moderation.} Our  staged pipeline achieves 86-100\% of full-model performance using only 6-57\% of training data (\autoref{tab:compare}).
Notably, we demonstrate this efficiency on challenging multi-class settings, whereas prior works tend to focus on simpler tasks \cite{guo2023investigation, kim2024label}. Unlike commercial APIs—where 16-shot ICL exhibits high variance ($\pm$0.10) and can degrade performance (\autoref{tab:icl_full})—SMARTER produces deterministic, explainable outputs suitable for production deployment at a fraction of the cost. When labeled data is scarce and explainability is required, SMARTER provides cost-efficient, transparent production deployment versus commercial APIs.

\paragraph{Cross-model refinement enables targeted improvement.} \textit{T5}'s superior cross-trained F1 scores (\autoref{fig:x_refine}) show that weaker models can adopt stronger models' reasoning patterns. Practitioners facing category-specific weaknesses (\autoref{fig:xmodel_anl}) can train multiple LLMs, then apply cross-model finetuning at high-quality checkpoints (e.g., $K=128$) to transfer strengths. Architectural diversity (encoder-decoder vs. decoder-only) appears beneficial, as \textit{Llama} enhances reasoning diversity that \textit{T5} leverages \cite{haollm}.

\paragraph{Human oversight remains essential.} Human evaluation (\autoref{tab:mturk_result}) shows category-specific preferences: \textit{Llama} excels on \textit{Normal} posts (3:1) while matching on \textit{Hate}. Practitioners should periodically validate explanations, especially after cross-model training, to maintain consistency (\autoref{tab:nli}).

We hope that SMARTER will enable practitioners to effectively and efficiently deploy explanation-based content moderation pipelines.

\section{Limitations}
We discuss several limitations in our research, along with their implications for future explorations.

\paragraph{Generalization beyond English} Our work focuses exclusively on English corpora. In reality, social media exists for every language. As both English-based LLMs and research literature often attract much more attention, we hope this framework will be adapted for other languages. 

\paragraph{Augmentation to base framework} In our experiments, samples are randomly sampled from the label space. However, more strategic sampling choices (e.g., selecting based on how the level of classification difficulty calibrated on validation set) may boost performance. We encourage practitioners to explore augmentation to maximize the utility of our framework. 

\paragraph{Model selection} While we perform extensive experiments, our selection of 2 models could be expanded. At the cost of more training resource, training more models with different architectures or generation offer extra options for performance comparison for mix-and-match cross-training. We encourage other researchers to experiment with more LLMs using our framework. 

\paragraph{Limited scope of human validation} Due to budget constraints, we are only able to perform human validation for 1 dataset. In practice, it is recommended to solicit a larger pool of annotators to ensure appropriate amount of human oversight. 

\paragraph{Risk of bias in automated explanation} While our framework aims to improve performance and versatility in content moderation, it does not alleviate the fundamental sensitive nature of this task. Content moderation could be misappropriated to suppress free speech or harm marginalized groups \cite{dias2020content, kozyreva2023resolving}. Other works have noted that LLMs exhibit bias behaviors in certain applications \cite{nguyen2023towards, gu2025large}. We urge researchers and practitioners to maintain vigilance when integrating our framework with human supervision to ensure ethical standards \cite{lai2022human,cao2024toxicity}.

\section{Ethical Considerations}
The authors are not aware of any ethical problems in the development of this work. This research uses publicly available dataset. We also further synonymize the content to the best of our ability to provide extra caution for privacy. We acknowledge that our framework and associated techniques could be abused for harmful purposes. 

\section{Acknowledgment}
This work is funded by the NSF under Grant No.
2229885 (NSF Institute for Trustworthy AI in Law
and Society, TRAILS). We thank the service of ACL
ARR reviewers, area chairs and the editors of the
ACL conference for our paper’s publication.

\bibliography{custom}

@article{hwang2015social,
  title={Social media as a tool for social movements: The effect of social media use and social capital on intention to participate in social movements},
  author={Hwang, Hyesun and Kim, Kee-Ok},
  journal={International Journal of Consumer Studies},
  volume={39},
  number={5},
  pages={478--488},
  year={2015},
  publisher={Wiley Online Library}
}

@article{cinelli2021echo,
  title={The echo chamber effect on social media},
  author={Cinelli, Matteo and De Francisci Morales, Gianmarco and Galeazzi, Alessandro and Quattrociocchi, Walter and Starnini, Michele},
  journal={Proceedings of the National Academy of Sciences},
  volume={118},
  number={9},
  pages={e2023301118},
  year={2021},
  publisher={National Acad Sciences}
}

@article{gelber2021differentiating,
  title={Differentiating hate speech: a systemic discrimination approach},
  author={Gelber, Katharine},
  journal={Critical Review of International Social and Political Philosophy},
  year={2021},
  publisher={Taylor \& Francis}
}

@article{anjum2024hate,
  title={Hate speech, toxicity detection in online social media: a recent survey of state of the art and opportunities},
  author={Anjum and Katarya, Rahul},
  journal={International Journal of Information Security},
  volume={23},
  number={1},
  pages={577--608},
  year={2024},
  publisher={Springer}
}

@article{castano2021internet,
  title={Internet, social media and online hate speech. Systematic review},
  author={Casta{\~n}o-Pulgar{\'\i}n, Sergio Andr{\'e}s and Su{\'a}rez-Betancur, Natalia and Vega, Luz Magnolia Tilano and L{\'o}pez, Harvey Mauricio Herrera},
  journal={Aggression and violent behavior},
  volume={58},
  pages={101608},
  year={2021},
  publisher={Elsevier}
}

@article{windisch2022online,
  title={Online interventions for reducing hate speech and cyberhate: A systematic review},
  author={Windisch, Steven and Wiedlitzka, Susann and Olaghere, Ajima and Jenaway, Elizabeth},
  journal={Campbell systematic reviews},
  volume={18},
  number={2},
  pages={e1243},
  year={2022},
  publisher={Wiley Online Library}
}

@article{tontodimamma2021thirty,
  title={Thirty years of research into hate speech: topics of interest and their evolution},
  author={Tontodimamma, Alice and Nissi, Eugenia and Sarra, Annalina and Fontanella, Lara},
  journal={Scientometrics},
  volume={126},
  pages={157--179},
  year={2021},
  publisher={Springer}
}

@article{baker2020covid19,
  title={<? covid19?> the challenges of responding to misinformation during a pandemic: content moderation and the limitations of the concept of harm},
  author={Baker, Stephanie Alice and Wade, Matthew and Walsh, Michael James},
  journal={Media International Australia},
  volume={177},
  number={1},
  pages={103--107},
  year={2020},
  publisher={SAGE Publications Sage UK: London, England}
}

@article{spence2023psychological,
  title={The psychological impacts of content moderation on content moderators: A qualitative study},
  author={Spence, Ruth and Bifulco, Antonia and Bradbury, Paula and Martellozzo, Elena and DeMarco, Jeffrey},
  journal={Cyberpsychology: Journal of Psychosocial Research on Cyberspace},
  volume={17},
  number={4},
  year={2023}
}

@inproceedings{d2020bert,
  title={Bert and fasttext embeddings for automatic detection of toxic speech},
  author={d'Sa, Ashwin Geet and Illina, Irina and Fohr, Dominique},
  booktitle={2020 International Multi-Conference on:“Organization of Knowledge and Advanced Technologies”(OCTA)},
  pages={1--5},
  year={2020},
  organization={IEEE}
}

@inproceedings{sumanth2022toxic,
  title={Toxic speech classification using machine learning algorithms},
  author={Sumanth, Pabba and Samiuddin, Syed and Jamal, K and Domakonda, Srikanth and Shivani, Pathi},
  booktitle={2022 International Conference on Electronic Systems and Intelligent Computing (ICESIC)},
  pages={257--263},
  year={2022},
  organization={IEEE}
}

@article{fortuna2021well,
  title={How well do hate speech, toxicity, abusive and offensive language classification models generalize across datasets?},
  author={Fortuna, Paula and Soler-Company, Juan and Wanner, Leo},
  journal={Information Processing \& Management},
  volume={58},
  number={3},
  pages={102524},
  year={2021},
  publisher={Elsevier}
}

@article{sap2021annotators,
  title={Annotators with attitudes: How annotator beliefs and identities bias toxic language detection},
  author={Sap, Maarten and Swayamdipta, Swabha and Vianna, Laura and Zhou, Xuhui and Choi, Yejin and Smith, Noah A},
  journal={arXiv preprint arXiv:2111.07997},
  year={2021}
}

@inproceedings{zhao2021comparative,
  title={A comparative study of using pre-trained language models for toxic comment classification},
  author={Zhao, Zhixue and Zhang, Ziqi and Hopfgartner, Frank},
  booktitle={Companion Proceedings of the Web Conference 2021},
  pages={500--507},
  year={2021}
}

@article{wei2021finetuned,
  title={Finetuned language models are zero-shot learners},
  author={Wei, Jason and Bosma, Maarten and Zhao, Vincent Y and Guu, Kelvin and Yu, Adams Wei and Lester, Brian and Du, Nan and Dai, Andrew M and Le, Quoc V},
  journal={arXiv preprint arXiv:2109.01652},
  year={2021}
}

@article{yao2024tree,
  title={Tree of thoughts: Deliberate problem solving with large language models},
  author={Yao, Shunyu and Yu, Dian and Zhao, Jeffrey and Shafran, Izhak and Griffiths, Tom and Cao, Yuan and Narasimhan, Karthik},
  journal={Advances in Neural Information Processing Systems},
  volume={36},
  year={2024}
}

@article{toraman2022large,
  title={Large-scale hate speech detection with cross-domain transfer},
  author={Toraman, Cagri and {\c{S}}ahinu{\c{c}}, Furkan and Yilmaz, Eyup Halit},
  journal={arXiv preprint arXiv:2203.01111},
  year={2022}
}

@inproceedings{bespalov2023towards,
  title={Towards Building a Robust Toxicity Predictor},
  author={Bespalov, Dmitriy and Bhabesh, Sourav and Xiang, Yi and Zhou, Liutong and Qi, Yanjun},
  booktitle={Proceedings of the 61st Annual Meeting of the Association for Computational Linguistics (Volume 5: Industry Track)},
  pages={581--598},
  year={2023}
}

@inproceedings{caselli2021hatebert,
  title={HateBERT: Retraining BERT for Abusive Language Detection in English},
  author={Caselli, Tommaso and Basile, Valerio and Mitrovi{\'c}, Jelena and Granitzer, Michael},
  booktitle={Proceedings of the 5th Workshop on Online Abuse and Harms (WOAH 2021)},
  pages={17--25},
  year={2021}
}

@inproceedings{sarkar2021fbert,
  title={fBERT: A Neural Transformer for Identifying Offensive Content},
  author={Sarkar, Diptanu and Zampieri, Marcos and Ranasinghe, Tharindu and Ororbia, Alexander},
  booktitle={Findings of the Association for Computational Linguistics: EMNLP 2021},
  pages={1792--1798},
  year={2021}
}

@inproceedings{nghiem2024define,
  title={“Define Your Terms”: Enhancing Efficient Offensive Speech Classification with Definition},
  author={Nghiem, Huy and Gupta, Umang and Morstatter, Fred},
  booktitle={Proceedings of the 18th Conference of the European Chapter of the Association for Computational Linguistics (Volume 1: Long Papers)},
  pages={1293--1309},
  year={2024}
}

@inproceedings{masud2024hate,
  title={Hate Personified: Investigating the role of LLMs in content moderation},
  author={Masud, Sarah and Singh, Sahajpreet and Hangya, Viktor and Fraser, Alexander and Chakraborty, Tanmoy},
  booktitle={Proceedings of the 2024 Conference on Empirical Methods in Natural Language Processing},
  pages={15847--15863},
  year={2024}
}

@article{jahan2024comprehensive,
  title={A Comprehensive Study on NLP Data Augmentation for Hate Speech Detection: Legacy Methods, BERT, and LLMs},
  author={Jahan, Md Saroar and Oussalah, Mourad and Beddia, Djamila Romaissa and Arhab, Nabil and others},
  journal={arXiv preprint arXiv:2404.00303},
  year={2024}
}

@inproceedings{antypas2023robust,
  title={Robust Hate Speech Detection in Social Media: A Cross-Dataset Empirical Evaluation},
  author={Antypas, Dimosthenis and Camacho-Collados, Jose},
  booktitle={The 7th Workshop on Online Abuse and Harms (WOAH)},
  pages={231--242},
  year={2023}
}

@inproceedings{roy2023probing,
  title={Probing LLMs for hate speech detection: strengths and vulnerabilities},
  author={Roy, Sarthak and Harshvardhan, Ashish and Mukherjee, Animesh and Saha, Punyajoy},
  booktitle={Findings of the Association for Computational Linguistics: EMNLP 2023},
  pages={6116--6128},
  year={2023}
}

@article{kumarage2024harnessing,
  title={Harnessing artificial intelligence to combat online hate: Exploring the challenges and opportunities of large language models in hate speech detection},
  author={Kumarage, Tharindu and Bhattacharjee, Amrita and Garland, Joshua},
  journal={arXiv preprint arXiv:2403.08035},
  year={2024}
}

@inproceedings{di2024explanation,
  title={Is explanation all you need? an expert survey on llm-generated explanations for abusive language detection},
  author={Di Bonaventura, Chiara and Siciliani, Lucia and Basile, Pierpaolo and Penuela, Albert Merono and McGillivray, Barbara},
  booktitle={Tenth Italian Conference on Computational Linguistics (CLiC-it 2024)},
  year={2024}
}

@inproceedings{yang2023hare,
  title={HARE: Explainable Hate Speech Detection with Step-by-Step Reasoning},
  author={Yang, Yongjin and Kim, Joonkee and Kim, Yujin and Ho, Namgyu and Thorne, James and Yun, Se-Young},
  booktitle={Findings of the Association for Computational Linguistics: EMNLP 2023},
  pages={5490--5505},
  year={2023}
}

@inproceedings{li2023synthetic,
  title={Synthetic Data Generation with Large Language Models for Text Classification: Potential and Limitations},
  author={Li, Zhuoyan and Zhu, Hangxiao and Lu, Zhuoran and Yin, Ming},
  booktitle={Proceedings of the 2023 Conference on Empirical Methods in Natural Language Processing},
  pages={10443--10461},
  year={2023}
}

@article{tang2023does,
  title={Does synthetic data generation of llms help clinical text mining?},
  author={Tang, Ruixiang and Han, Xiaotian and Jiang, Xiaoqian and Hu, Xia},
  journal={arXiv preprint arXiv:2303.04360},
  year={2023}
}

@inproceedings{mathew2021hatexplain,
  title={Hatexplain: A benchmark dataset for explainable hate speech detection},
  author={Mathew, Binny and Saha, Punyajoy and Yimam, Seid Muhie and Biemann, Chris and Goyal, Pawan and Mukherjee, Animesh},
  booktitle={Proceedings of the AAAI conference on artificial intelligence},
  volume={35},
  number={17},
  pages={14867--14875},
  year={2021}
}

@inproceedings{elsherief2021latent,
  title={Latent Hatred: A Benchmark for Understanding Implicit Hate Speech},
  author={ElSherief, Mai and Ziems, Caleb and Muchlinski, David and Anupindi, Vaishnavi and Seybolt, Jordyn and De Choudhury, Munmun and Yang, Diyi},
  booktitle={Proceedings of the 2021 Conference on Empirical Methods in Natural Language Processing},
  pages={345--363},
  year={2021}
}

@inproceedings{nghiem-daume-iii-2024-hatecot,
    title = "{H}ate{COT}: An Explanation-Enhanced Dataset for Generalizable Offensive Speech Detection via Large Language Models",
    author = "Nghiem, Huy  and
      Daum{\'e}, Hal",
    editor = "Al-Onaizan, Yaser  and
      Bansal, Mohit  and
      Chen, Yun-Nung",
    booktitle = "Findings of the Association for Computational Linguistics: EMNLP 2024",
    month = nov,
    year = "2024",
    address = "Miami, Florida, USA",
    publisher = "Association for Computational Linguistics",
    url = "https://aclanthology.org/2024.findings-emnlp.343/",
    doi = "10.18653/v1/2024.findings-emnlp.343",
    pages = "5938--5956",

}

@inproceedings{kim2023cot,
  title={The CoT Collection: Improving Zero-shot and Few-shot Learning of Language Models via Chain-of-Thought Fine-Tuning},
  author={Kim, Seungone and Joo, Se and Kim, Doyoung and Jang, Joel and Ye, Seonghyeon and Shin, Jamin and Seo, Minjoon},
  booktitle={Proceedings of the 2023 Conference on Empirical Methods in Natural Language Processing},
  pages={12685--12708},
  year={2023}
}

@inproceedings{longpre2023flan,
  title={The flan collection: Designing data and methods for effective instruction tuning},
  author={Longpre, Shayne and Hou, Le and Vu, Tu and Webson, Albert and Chung, Hyung Won and Tay, Yi and Zhou, Denny and Le, Quoc V and Zoph, Barret and Wei, Jason and others},
  booktitle={International Conference on Machine Learning},
  pages={22631--22648},
  year={2023},
  organization={PMLR}
}

@article{dubey2024llama,
  title={The llama 3 herd of models},
  author={Dubey, Abhimanyu and Jauhri, Abhinav and Pandey, Abhinav and Kadian, Abhishek and Al-Dahle, Ahmad and Letman, Aiesha and Mathur, Akhil and Schelten, Alan and Yang, Amy and Fan, Angela and others},
  journal={arXiv preprint arXiv:2407.21783},
  year={2024}
}

@article{wei2022chain,
  title={Chain-of-thought prompting elicits reasoning in large language models},
  author={Wei, Jason and Wang, Xuezhi and Schuurmans, Dale and Bosma, Maarten and Xia, Fei and Chi, Ed and Le, Quoc V and Zhou, Denny and others},
  journal={Advances in neural information processing systems},
  volume={35},
  pages={24824--24837},
  year={2022}
}

@inproceedings{hulora,
  title={LoRA: Low-Rank Adaptation of Large Language Models},
  author={Hu, Edward J and Wallis, Phillip and Allen-Zhu, Zeyuan and Li, Yuanzhi and Wang, Shean and Wang, Lu and Chen, Weizhu and others},
  booktitle={International Conference on Learning Representations}, 
  year={2021}
}

@article{ethayarajh2024kto,
  title={Kto: Model alignment as prospect theoretic optimization},
  author={Ethayarajh, Kawin and Xu, Winnie and Muennighoff, Niklas and Jurafsky, Dan and Kiela, Douwe},
  journal={arXiv preprint arXiv:2402.01306},
  year={2024}
}

@article{rafailov2024direct,
  title={Direct preference optimization: Your language model is secretly a reward model},
  author={Rafailov, Rafael and Sharma, Archit and Mitchell, Eric and Manning, Christopher D and Ermon, Stefano and Finn, Chelsea},
  journal={Advances in Neural Information Processing Systems},
  volume={36},
  year={2024}
}

@article{christen2023review,
  title={A review of the F-measure: its history, properties, criticism, and alternatives},
  author={Christen, Peter and Hand, David J and Kirielle, Nishadi},
  journal={ACM Computing Surveys},
  volume={56},
  number={3},
  pages={1--24},
  year={2023},
  publisher={ACM New York, NY}
}

@inproceedings{devlin2019bert,
  title={Bert: Pre-training of deep bidirectional transformers for language understanding},
  author={Devlin, Jacob and Chang, Ming-Wei and Lee, Kenton and Toutanova, Kristina},
  booktitle={Proceedings of the 2019 conference of the North American chapter of the association for computational linguistics: human language technologies, volume 1 (long and short papers)},
  pages={4171--4186},
  year={2019}
}

@article{kozyreva2023resolving,
  title={Resolving content moderation dilemmas between free speech and harmful misinformation},
  author={Kozyreva, Anastasia and Herzog, Stefan M and Lewandowsky, Stephan and Hertwig, Ralph and Lorenz-Spreen, Philipp and Leiser, Mark and Reifler, Jason},
  journal={Proceedings of the National Academy of Sciences},
  volume={120},
  number={7},
  pages={e2210666120},
  year={2023},
  publisher={National Academy of Sciences}
}

@article{dias2020content,
  title={Content moderation technologies: Applying human rights standards to protect freedom of expression},
  author={Dias Oliva, Thiago},
  journal={Human Rights Law Review},
  volume={20},
  number={4},
  pages={607--640},
  year={2020},
  publisher={Oxford University Press}
}

@inproceedings{nguyen2023towards,
  title={Towards Conceptualization of “Fair Explanation”: Disparate Impacts of anti-Asian Hate Speech Explanations on Content Moderators},
  author={Nguyen, Tin and Xu, Jiannan and Roy, Aayushi and Daum{\'e} III, Hal and Carpuat, Marine},
  booktitle={Proceedings of the 2023 Conference on Empirical Methods in Natural Language Processing},
  pages={9696--9717},
  year={2023}
}

@article{gu2025large,
  title={Large Language Models Are Effective Human Annotation Assistants, But Not Good Independent Annotators},
  author={Gu, Feng and Li, Zongxia and Colon, Carlos Rafael and Evans, Benjamin and Mondal, Ishani and Boyd-Graber, Jordan Lee},
  journal={arXiv preprint arXiv:2503.06778},
  year={2025},
  url={https://arxiv.org/abs/2503.06778}
}

@inproceedings{lai2022human,
  title={Human-ai collaboration via conditional delegation: A case study of content moderation},
  author={Lai, Vivian and Carton, Samuel and Bhatnagar, Rajat and Liao, Q Vera and Zhang, Yunfeng and Tan, Chenhao},
  booktitle={Proceedings of the 2022 CHI Conference on Human Factors in Computing Systems},
  pages={1--18},
  year={2022}
}

@inproceedings{cao2024toxicity,
  title={Toxicity Detection is NOT all you Need: Measuring the Gaps to Supporting Volunteer Content Moderators through a User-Centric Method},
  author={Cao, Yang and Domingo, Lovely-Frances and Gilbert, Sarah and Mazurek, Michelle and Shilton, Katie and Iii, Hal Daum{\'e}},
  booktitle={Proceedings of the 2024 Conference on Empirical Methods in Natural Language Processing},
  pages={3567--3587},
  year={2024}
}

@inproceedings{haollm,
  title={LLM Reasoners: New Evaluation, Library, and Analysis of Step-by-Step Reasoning with Large Language Models},
  author={Hao, Shibo and Gu, Yi and Luo, Haotian and Liu, Tianyang and Shao, Xiyan and Wang, Xinyuan and Xie, Shuhua and Ma, Haodi and Samavedhi, Adithya and Gao, Qiyue and others},
  booktitle={First Conference on Language Modeling}, 
  year={2024}
}

@incollection{kahneman2013prospect,
  title={Prospect theory: An analysis of decision under risk},
  author={Kahneman, Daniel and Tversky, Amos},
  booktitle={Handbook of the fundamentals of financial decision making: Part I},
  pages={99--127},
  year={2013},
  publisher={World Scientific}
}

@article{warner2024smarter,
  title={Smarter, better, faster, longer: A modern bidirectional encoder for fast, memory efficient, and long context finetuning and inference},
  author={Warner, Benjamin and Chaffin, Antoine and Clavi{\'e}, Benjamin and Weller, Orion and Hallstr{\"o}m, Oskar and Taghadouini, Said and Gallagher, Alexis and Biswas, Raja and Ladhak, Faisal and Aarsen, Tom and others},
  journal={arXiv preprint arXiv:2412.13663},
  year={2024}
}

@article{rad2025refining,
  title={Refining Input Guardrails: Enhancing LLM-as-a-Judge Efficiency Through Chain-of-Thought Fine-Tuning and Alignment},
  author={Rad, Melissa Kazemi and Nghiem, Huy and Luo, Andy and Wadhwa, Sahil and Sorower, Mohammad and Rawls, Stephen},
  journal={arXiv preprint arXiv:2501.13080},
  year={2025}
}

@article{nguyen2024decade,
  title={A Decade of Tweets: Visualizing Racial Sentiments Towards Minoritized Groups in the United States Between 2011 and 2021},
  author={Nguyen, Thu T and Merchant, Junaid S and Yue, Xiaohe and Mane, Heran and Wei, Hanxue and Huang, Dina and Gowda, Krishik N and Makres, Katrina and Najib, Crystal and Nghiem, Huy T and others},
  journal={Epidemiology},
  volume={35},
  number={1},
  pages={51--59},
  year={2024},
  publisher={LWW}
}

@article{ouyang2022training,
  title={Training language models to follow instructions with human feedback},
  author={Ouyang, Long and Wu, Jeffrey and Jiang, Xu and Almeida, Diogo and Wainwright, Carroll and Mishkin, Pamela and Zhang, Chong and Agarwal, Sandhini and Slama, Katarina and Ray, Alex and others},
  journal={Advances in neural information processing systems},
  volume={35},
  pages={27730--27744},
  year={2022}
}

@article{shao2024deepseekmath,
  title={Deepseekmath: Pushing the limits of mathematical reasoning in open language models},
  author={Shao, Zhihong and Wang, Peiyi and Zhu, Qihao and Xu, Runxin and Song, Junxiao and Bi, Xiao and Zhang, Haowei and Zhang, Mingchuan and Li, YK and others},
  journal={arXiv preprint arXiv:2402.03300},
  year={2024}
}

@article{zweiger2025self,
  title={Self-Adapting Language Models},
  author={Zweiger, Adam and Pari, Jyothish and Guo, Han and Aky{\"u}rek, Ekin and Kim, Yoon and Agrawal, Pulkit},
  journal={arXiv preprint arXiv:2506.10943},
  year={2025}
}

@article{madaan2023self,
  title={Self-refine: Iterative refinement with self-feedback},
  author={Madaan, Aman and Tandon, Niket and Gupta, Prakhar and Hallinan, Skyler and Gao, Luyu and Wiegreffe, Sarah and Alon, Uri and Dziri, Nouha and Prabhumoye, Shrimai and Yang, Yiming and others},
  journal={Advances in Neural Information Processing Systems},
  volume={36},
  pages={46534--46594},
  year={2023}
}

@article{wang2022self,
  title={Self-instruct: Aligning language models with self-generated instructions},
  author={Wang, Yizhong and Kordi, Yeganeh and Mishra, Swaroop and Liu, Alisa and Smith, Noah A and Khashabi, Daniel and Hajishirzi, Hannaneh},
  journal={arXiv preprint arXiv:2212.10560},
  year={2022}
}

@inproceedings{calabrese2024explainability,
  title={Explainability and hate speech: Structured explanations make social media moderators faster},
  author={Calabrese, Agostina and Neves, Leonardo and Shah, Neil and Bos, Maarten and Ross, Bj{\"o}rn and Lapata, Mirella and Barbieri, Francesco},
  booktitle={Proceedings of the 62nd Annual Meeting of the Association for Computational Linguistics (Volume 2: Short Papers)},
  pages={398--408},
  year={2024}
}

@inproceedings{wang2025sahsd,
  title={SAHSD: Enhancing Hate Speech Detection in LLM-Powered Web Applications via Sentiment Analysis and Few-Shot Learning},
  author={Wang, Yulong and Li, Hong and Wei, Ni},
  booktitle={Proceedings of the ACM on Web Conference 2025},
  pages={3014--3025},
  year={2025}
}

@article{almohaimeed2025towards,
  title={Towards Generalizable Generic Harmful Speech Datasets for Implicit Hate Speech Detection},
  author={Almohaimeed, Saad and Almohaimeed, Saleh and Turgut, Damla and B{\"o}l{\"o}ni, Ladislau},
  journal={arXiv preprint arXiv:2506.16476},
  year={2025}
}

@inproceedings{negru2025morphnli,
  title={MorphNLI: A Stepwise Approach to Natural Language Inference Using Text Morphing},
  author={Negru, Vlad Andrei and Vacareanu, Robert and Lemnaru, Camelia and Surdeanu, Mihai and Potolea, Rodica},
  booktitle={Findings of the Association for Computational Linguistics: NAACL 2025},
  pages={6938--6953},
  year={2025}
}

@inproceedings{gallipoli2025not,
  title={It is not a piece of cake for GPT: Explaining Textual Entailment Recognition in the presence of Figurative Language},
  author={Gallipoli, Giuseppe and Cagliero, Luca},
  booktitle={Proceedings of the 31st International Conference on Computational Linguistics},
  pages={9656--9674},
  year={2025}
}

@inproceedings{guo2023investigation,
  title={An investigation of large language models for real-world hate speech detection},
  author={Guo, Keyan and Hu, Alexander and Mu, Jaden and Shi, Ziheng and Zhao, Ziming and Vishwamitra, Nishant and Hu, Hongxin},
  booktitle={2023 International Conference on Machine Learning and Applications (ICMLA)},
  pages={1568--1573},
  year={2023},
  organization={IEEE}
}

@inproceedings{kim2024label,
  title={Label-aware Hard Negative Sampling Strategies with Momentum Contrastive Learning for Implicit Hate Speech Detection},
  author={Kim, Jaehoon and Jin, Seungwan and Park, Sohyun and Park, Someen and Han, Kyungsik},
  booktitle={Findings of the Association for Computational Linguistics ACL 2024},
  pages={16177--16188},
  year={2024}
}

@inproceedings{nghiem2025balancing,
  title={Balancing Safety and Helpfulness in Healthcare AI Assistants through Iterative Preference Alignment},
  author={Nghiem, Huy and Panda, S. and Khatwani, D. and Nguyen, H. V. and Kenthapadi, K. and Nguyen, H. D.},
  booktitle={ML4H Symposium},
  year={2025},
  url={https://arxiv.org/abs/2512.04210},
  note={arXiv preprint arXiv:2512.04210}
}

\clearpage
\appendix
\section*{Appendix}
\label{sec:appendix}

\section{Amazon Mechanical Turk Annotation}
We obtained IRB approval before soliciting annotation from the crowd workers. 14 workers were recruited to annotate the 342 explanations introduced in Section \ref{sec:human_val}. Samples are assigned at random by the AMT platform to each annotator. Due to budget constraint, we opt to collect annotation for \textit{HateXplain} to preserve statistical power instead of spreading across all 3 datasets. The breakdown of the annotators' demographic is below:
\begin{itemize}
    \item \textbf{Gender}: Female (10), Male (3), Non-binary (1) 
    \item \textbf{Age}: 18-29 (2), 30-39 (5), 40-49 (5), 50+ (2)
    \item \textbf{Race/Ethnicity}: Asian (3), Black (2), Hispanic/Latino (1), White (8)
    \item \textbf{Country of residence}: United States (14)
    \item \textbf{Education level}: 2-year college or equivalent (4), 4-year college or equivalent (6), High school or equivalent (3), Master degree or above (1)
\end{itemize}

\begin{figure*}[h]
  \centering
  \includegraphics[width=0.9\textwidth]{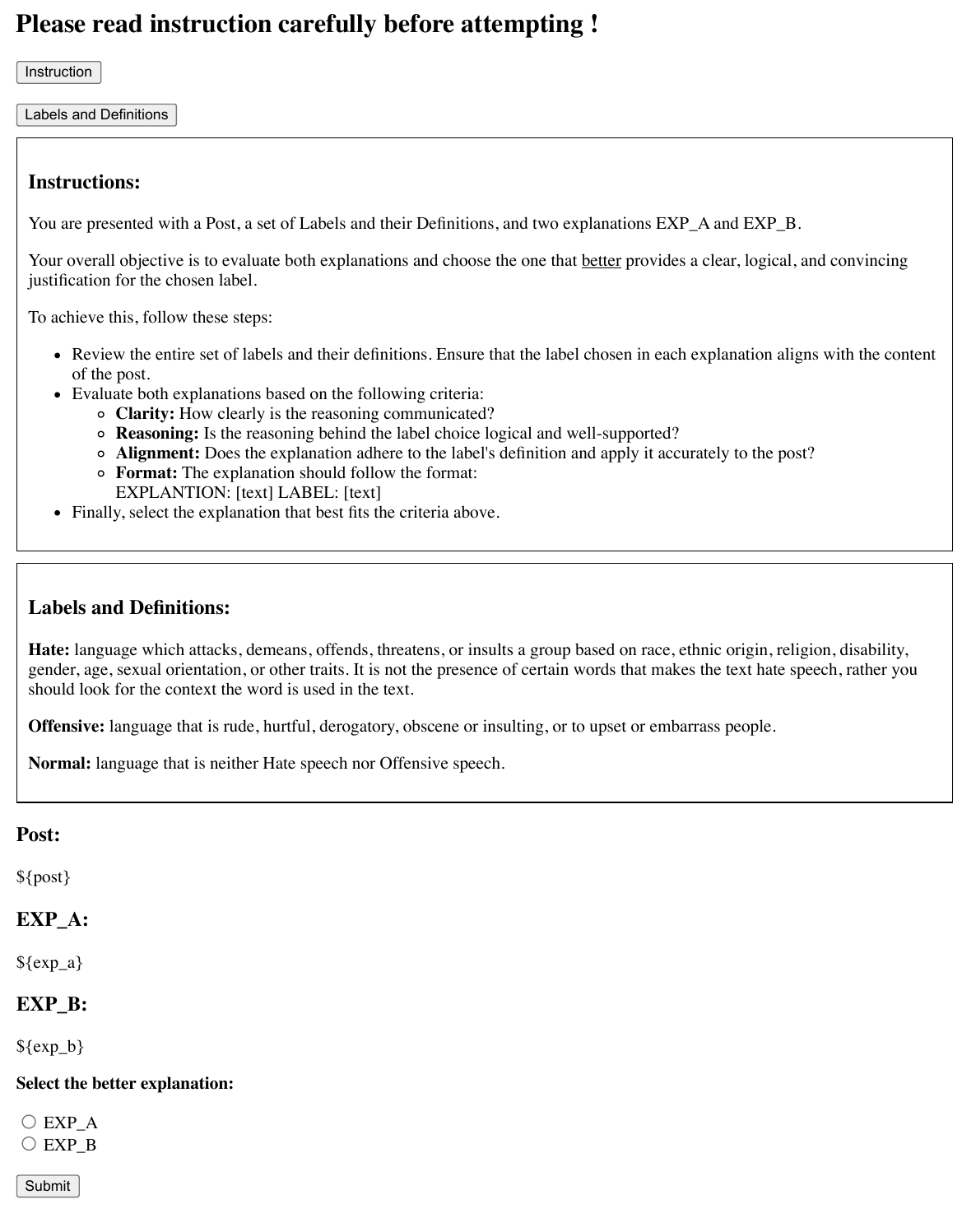}
  \caption{Template to collect annotation for preference on explanations for Amazon Mechanical Turks crowdworkers.}
  \label{fig:mturk}
\end{figure*}

\section{Data Pre-processing}
\label{apx:data}
The datasets used in this work are open-source. We peruse them as research artifacts accordingly to their license. To further mitigate risks of privacy, we anonymize posts by replacing  user handles with the string \textit{<user>} and remove URLs if they appear in the context of the posts.

We obtain the seed explanations from the repository associated with \citet{nghiem-daume-iii-2024-hatecot} and merge them with the correct identifier keys.

\begin{table*}[t]
\centering
\footnotesize
\renewcommand{\arraystretch}{1.35}  
\newcolumntype{C}[1]{>{\centering\arraybackslash}p{#1}}
\newcolumntype{R}[1]{>{\raggedleft\arraybackslash}p{#1}} 
{\begin{tabular}{p{2cm}R{1cm}R{0.75cm}R{0.5cm}R{0.2cm}p{1cm}p{7cm}}
\toprule  
\textbf{Dataset} & \textbf{Total Size} & \textbf{Split Ratio} & \textbf{K val} & \textbf{K test} & \textbf{Platform} & \textbf{Label Space} \\
\midrule  
\citetalias{mathew2021hatexplain} & 20,148 & 60:20:20 & 50 & 400 & G, T & Normal, Offensive, Hate                \\
\citetalias{elsherief2021latent} & 19,112 & 60:20:20 & 50 & 400 & T    & Not Hate, Explicit Hate, Implicit Hate \\
Implicit Hate & 4,153 & 50:20:30 & 50 & 150 & T & White Grievance, Incitement to Violence, Inferiority Language, Irony, Stereotypes and Misinformation, Threatening and Intimidation \\
\hline
\end{tabular}}
\caption{Datasets for evaluating our framework. \textit{Split Ratio} designates the proportion of the train:validation:test split. \textit{K val} and \textit{K test} are the number of data points per class to perform $K$-shot sampling from the Validation and Test set. For \textit{Platform}, \textit{G} denotes Gab, \textit{T} for X (formerly Twitter).   }
\label{tab:test_data}
\end{table*}

\section{Obtaining Explanations for In-domain Data}
The prompt template in \autoref{fig:cond_prompt} is used to collect explanation conditioned on a single label. Note that more labels (and their definitions) may also be added to this section as desired.

\begin{table}[t]
\footnotesize
\centering
\setlength{\tabcolsep}{10pt}
\renewcommand{\arraystretch}{1.1}
\begin{tabular}{llr}
\toprule
\textbf{Dataset} & \textbf{Model} & \textbf{Macro F1} \\
\midrule
\textit{HateXplain} & Llama\_OTS & 0.52 \\
                    & T5\_OTS    & 0.56 \\
\midrule
\textit{Latent Hate} & Llama\_OTS & 0.53 \\
                     & T5\_OTS    & 0.42 \\
\midrule
\textit{Implicit Hate} & Llama\_OTS & 0.32 \\
                       & T5\_OTS    & 0.33 \\
\bottomrule
\end{tabular}
\caption{Comparison of off-the-shelf (OTS) model performance across datasets. LLMs are prompted to produce answers without explanations in this setting as shown in \autoref{fig:cls_prompt_noexp}.}
\label{tab:ots_f1}
\end{table}

\begin{figure*}[!h]
    \centering
    \fbox{\colorbox{lightgreen}{\parbox{0.95\textwidth}{%
            \ttfamily
\#\#\# \textbf{Instruction}:\\
By considering their corresponding Definitions, label the following post with only one of these categories: \{categories\}.\\[4pt]
Provide your response in the following format:\\
LABEL: [text]\\[6pt]
\#\#\# \textbf{Definitions}:\\
\{definitions\}\\[4pt]
\#\#\# \textbf{Post}:\\
\{post\}\\[4pt]
\#\#\# \textbf{Response}: [model output]
}}}
    \caption{Prompt template for classification tasks \textbf{without explanation generation}. The model is instructed to directly output a label based on the provided definitions.}
    \label{fig:cls_prompt_noexp}
\end{figure*}

\begin{table}[!h]
    \centering
    \footnotesize
    \resizebox{\columnwidth}{!}{%
    \begin{tabular}{l l c c c c}
        \hline
        \textbf{Dataset} & \textbf{Model} & \textbf{Min} & \textbf{Mean} & \textbf{Max} & \textbf{Range} \\
        \hline
        
        \multirow{4}{*}{\textit{HateXplain}} 
            & GPT-4o-mini     & 0.22 & 0.29 & 0.42 & 0.20 \\
            & GPT-4.1         & 0.47 & 0.52 & 0.60 & 0.13 \\
            & GPT-5-chat      & 0.61 & 0.62 & 0.63 & 0.02 \\
            & Qwen-32B        & 0.52 & 0.55 & 0.59 & 0.07 \\
        \hline
        
        \multirow{4}{*}{\textit{Latent Hate}} 
            & GPT-4o-mini     & 0.17 & 0.25 & 0.34 & 0.17 \\
            & GPT-4.1         & 0.58 & 0.63 & 0.67 & 0.09 \\
            & GPT-5-chat      & 0.54 & 0.60 & 0.65 & 0.11 \\
            & Qwen-32B        & 0.54 & 0.57 & 0.60 & 0.06 \\
        \hline
        
        \multirow{4}{*}{\textit{Implicit Hate}} 
            & GPT-4o-mini     & 0.14 & 0.15 & 0.16 & 0.02 \\
            & GPT-4.1         & 0.30 & 0.38 & 0.51 & 0.21 \\
            & GPT-5-chat      & 0.29 & 0.40 & 0.51 & 0.22 \\
            & Qwen-32B        & 0.44 & 0.48 & 0.51 & 0.07 \\
        \hline
    \end{tabular}
    }
    \caption{Detailed 16-shot in-context learning (ICL) results for commercial models across 3 random seeds. For each dataset, we report the minimum, mean, and maximum Macro F1 scores, along with the range (Max -- Min) to illustrate performance variability. Models were prompted with 16 examples per class. High variance, particularly for GPT-4o-mini and on complex tasks like Implicit Hate, demonstrates ICL's brittleness for nuanced multi-class toxicity detection.}
    \label{tab:icl_full}
\end{table}

\label{apx:get_exp}

\begin{figure*}[!h]
    \centering
    \fbox{\colorbox{lightgreen}{\parbox{0.95\textwidth}{%
            \ttfamily
\#\#\# \textbf{Instruction}:\\
By considering their corresponding Definitions, label the following post with only one of these categories: {categories}.
Provide a brief paragraph to explain step-by-step why the post should be classified with the provided Label based on the given Definitions.\\
If this post targets a group or entity relevant to the definition of the specified Label, explain who this target is/are and how that justifies the chosen Label. \\
Provide your message in the following format: \\
EXPLANATION: [text] \\
LABEL: [text] \\
\#\#\# \textbf{Definitions}: \\
Hate: Language that targets [...] \\
Abusive: Language that is [...] \\
Normal: Does not contain [...] \\
\#\#\# \textbf{Post}: \\
\{post\} \\
\#\#\# \textbf{Response}: [model output]
}}}
    \caption{Prompt template for classification tasks. The definition block should contain the set's full label space}
    \label{fig:cls_prompt}
\end{figure*}

\begin{figure*}[!h]
    \centering
    \fbox{\colorbox{lightgreen}{\parbox{0.95\textwidth}{%
            \ttfamily
\#\#\# \textbf{Instruction}:\\
By considering the following Post, the Label and its Definition, provide a brief paragraph to explain step-by-step why the post should be classified with the provided Label based on the given Definition.\\
If this post targets a group or entity relevant to the definition of the specified Label, explain who this target is/are and how that justifies the chosen Label.\\
Provide your message in the following format: \\
EXPLANATION: [text] \\
LABEL: [text] \\
\#\#\# \textbf{Definitions}: \\
Hate: Language that targets [...] \\
\#\#\# Post: \\
\{post\} \\
\#\#\# \textbf{Response}:
}}}
    \caption{Prompt template to obtain explanation conditioned on a single label and its definition.}
    \label{fig:cond_prompt}
\end{figure*}

\begin{figure*}[!h]
    \centering
    \fbox{\colorbox{lightgreen}{\parbox{0.95\textwidth}{%
            \ttfamily
\#\#\# \textbf{Instruction}:\\
You are given a model's prediction and its explanation for a social media post.\\
Your task is to judge whether the explanation is logically consistent with the predicted label.\\[4pt]
Possible outputs: \\
- ENTAIL: The explanation supports and is faithful to the predicted label.\\
- CONTRADICT: The explanation contradicts the predicted label.\\
- NEUTRAL: The explanation is unrelated, vague, or does not clearly support the predicted label.\\
- UNDEFINED: Either the predicted label or the explanation is missing or empty.\\[4pt]
Return only one word: ENTAIL, CONTRADICT, NEUTRAL, or UNDEFINED.\\[6pt]
\#\#\# \textbf{Predicted Label}:\\
\{pred\_label\}\\[4pt]
\#\#\# \textbf{Model Explanation}:\\
\{explanation\}\\[4pt]
\#\#\# \textbf{Response}: [model output]
}}}
    \caption{Prompt template for evaluating \textbf{explanation–label consistency}. The model is asked to judge whether a generated explanation is logically consistent with its predicted label.}
    \label{fig:nli_prompt_1}
\end{figure*}

\begin{figure*}[!h]
    \centering
    \fbox{\colorbox{lightgreen}{\parbox{0.95\textwidth}{%
            \ttfamily
\#\#\# \textbf{Instruction}:\\
You are given a model's predicted label and its corresponding explanation.\\
Your task is to judge whether the explanation is logically consistent with the definition of that label.\\[4pt]
Possible outputs: \\
- ENTAIL: The explanation clearly aligns with and supports the definition.\\
- CONTRADICT: The explanation conflicts with the definition.\\
- NEUTRAL: The explanation is vague, unrelated, or does not clearly support the definition.\\[4pt]
Return only one word: ENTAIL, CONTRADICT, or NEUTRAL.\\[6pt]
\#\#\# \textbf{Official Definition}:\\
\{definition\}\\[4pt]
\#\#\# \textbf{Model Explanation}:\\
\{explanation\}\\[4pt]
\#\#\# \textbf{Response}: [model output]
}}}
    \caption{Prompt template for evaluating \textbf{explanation–definition consistency}. The model is asked to assess whether a generated explanation logically aligns with the formal definition of the predicted label.}
    \label{fig:nli_prompt_2}
\end{figure*}

\section{Stylistic Analysis}
\begin{figure*}[h]
    \centering
    \begin{subfigure}{\linewidth}
        \centering
        \includegraphics[width=\linewidth]{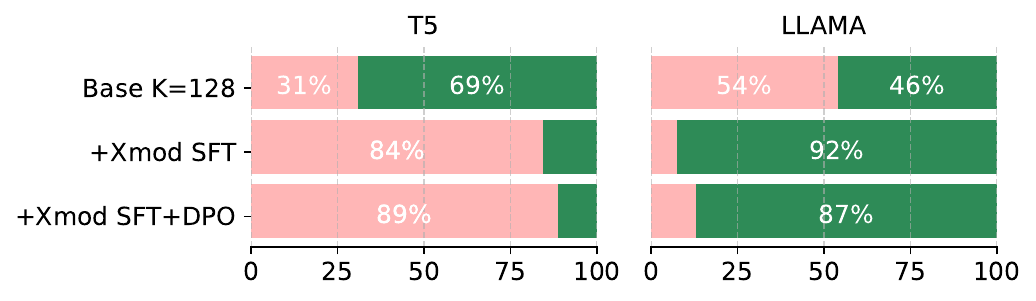}
        \caption{HateXplain}
        \label{fig:bert_hatexplain}
    \end{subfigure}

    \begin{subfigure}{\linewidth}
        \centering
        \includegraphics[width=\linewidth]{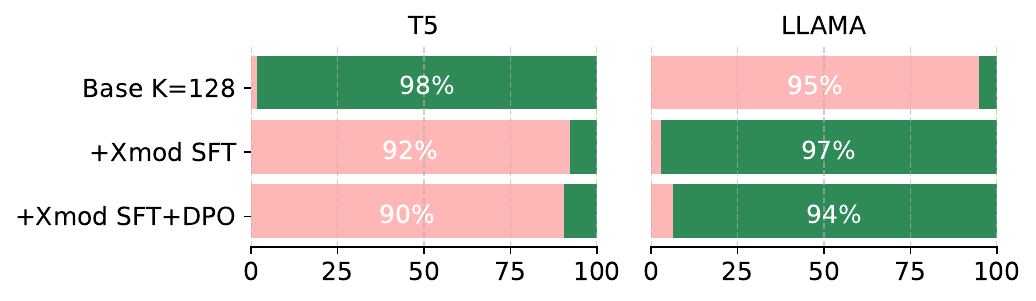}
        \caption{Latent Hate}
        \label{fig:bert_latent}
    \end{subfigure}

    \begin{subfigure}{\linewidth}
        \centering
        \includegraphics[width=\linewidth]{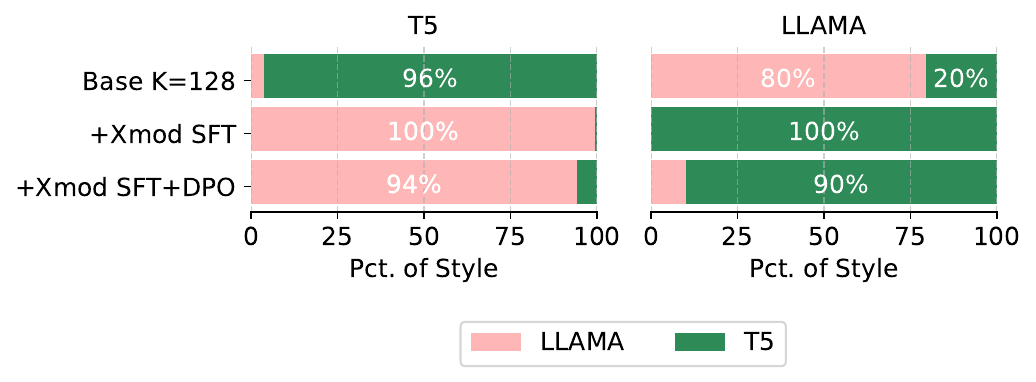}
        \caption{Implicit Hate}
        \label{fig:bert_implicit}
    \end{subfigure}

\caption{Percentage distribution of \textit{Llama}- and \textit{T5}-style explanations on test sets  by  BERT style classifiers, pre (\textit{Base}) and post cross-model (\textit{Xmod}) refined. }
\label{fig:bert_datasets}
\end{figure*}

In \autoref{fig:bert_datasets}, we show the distribution of \textit{Llama} vs \textit{T5} style pre- and post cross-model refined as recognized by the BERT style classifier.

\begin{figure*}[ht]
    \centering

    \begin{subfigure}[b]{\textwidth}
        \centering
        \includegraphics[width=0.48\linewidth]{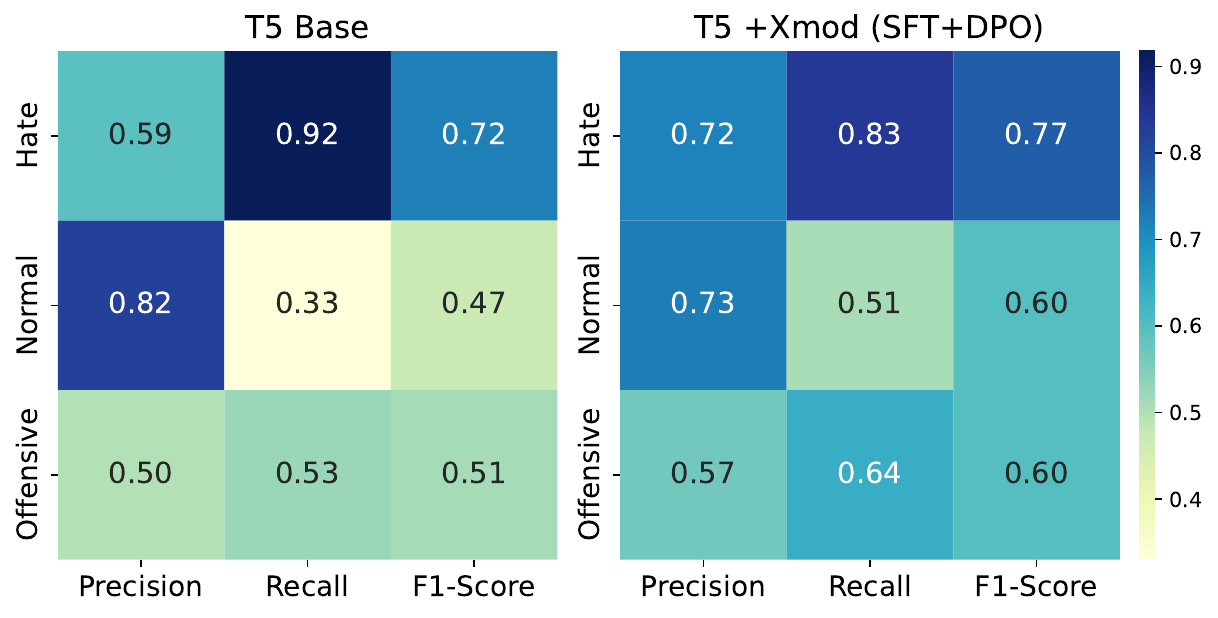}
        \includegraphics[width=0.48\linewidth]{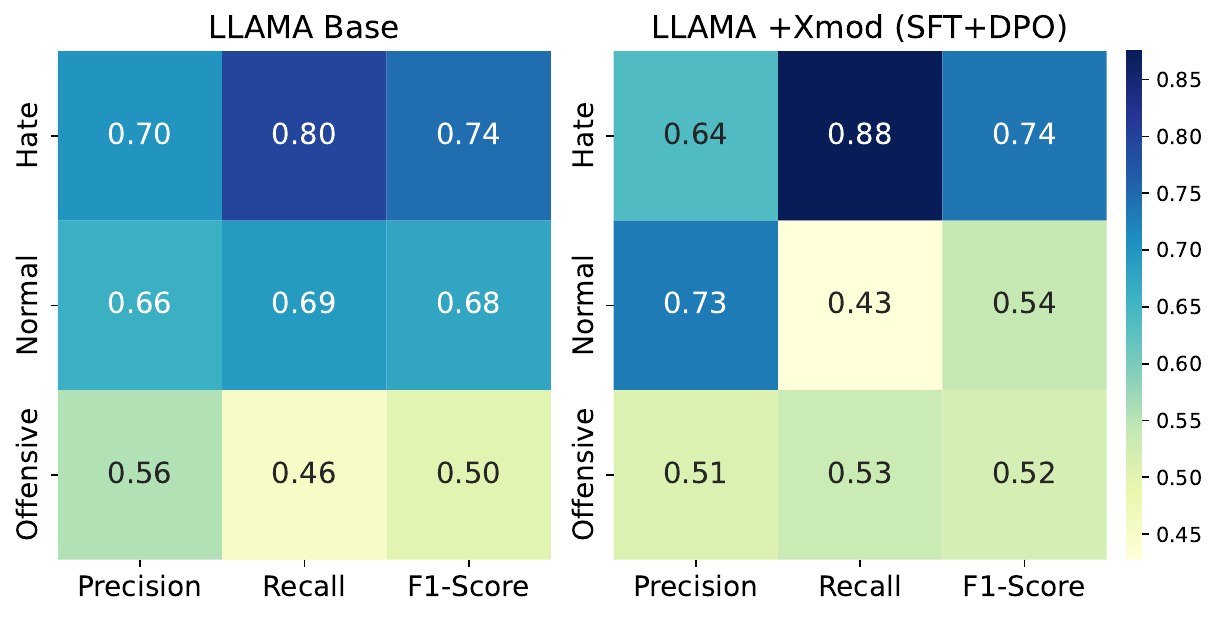}
        \caption{\textit{HateXlain}}
        \label{fig:hatexplain_error}
    \end{subfigure}
    \hfill

    \begin{subfigure}[b]{\textwidth}
        \centering
        \includegraphics[width=0.48\linewidth]{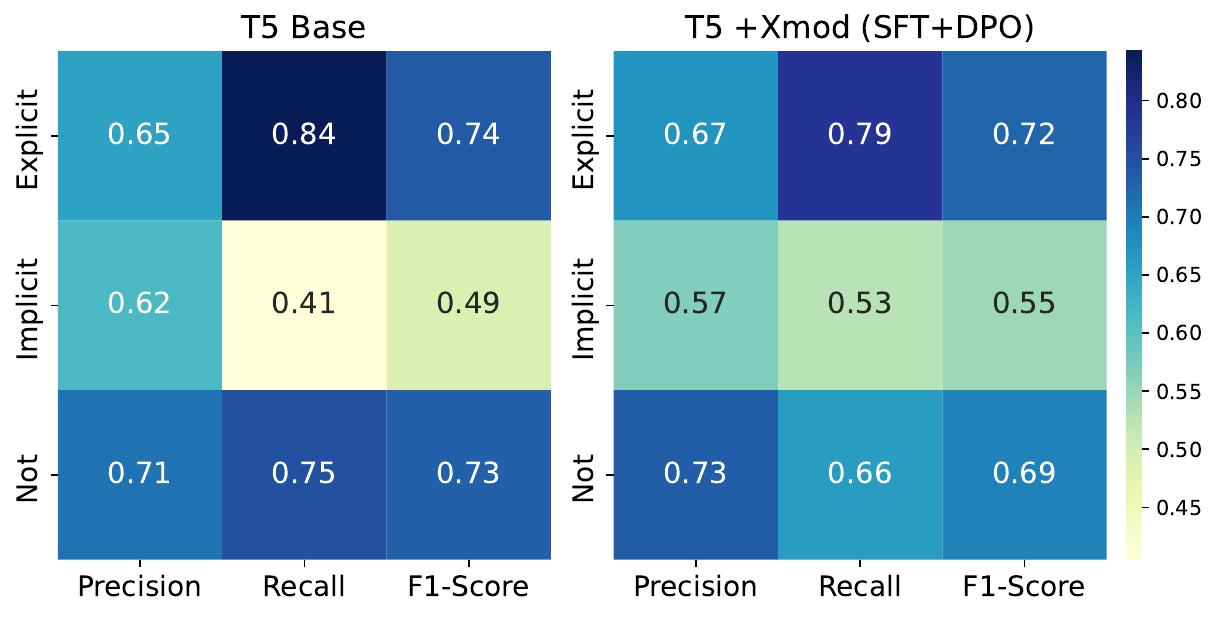}
        \includegraphics[width=0.48\linewidth]{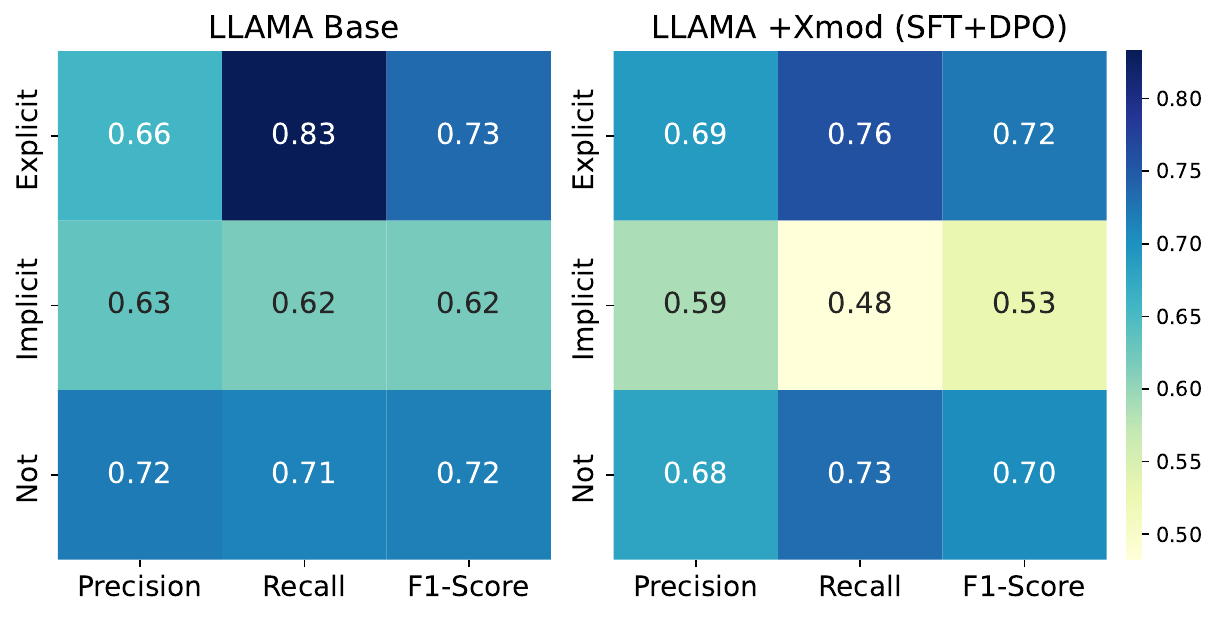}
        \caption{\textit{Latent Hate}}
        \label{fig:latent_error}
    \end{subfigure}
    \hfill

    \begin{subfigure}[b]{\textwidth}
        \centering
        \includegraphics[width=0.48\linewidth]{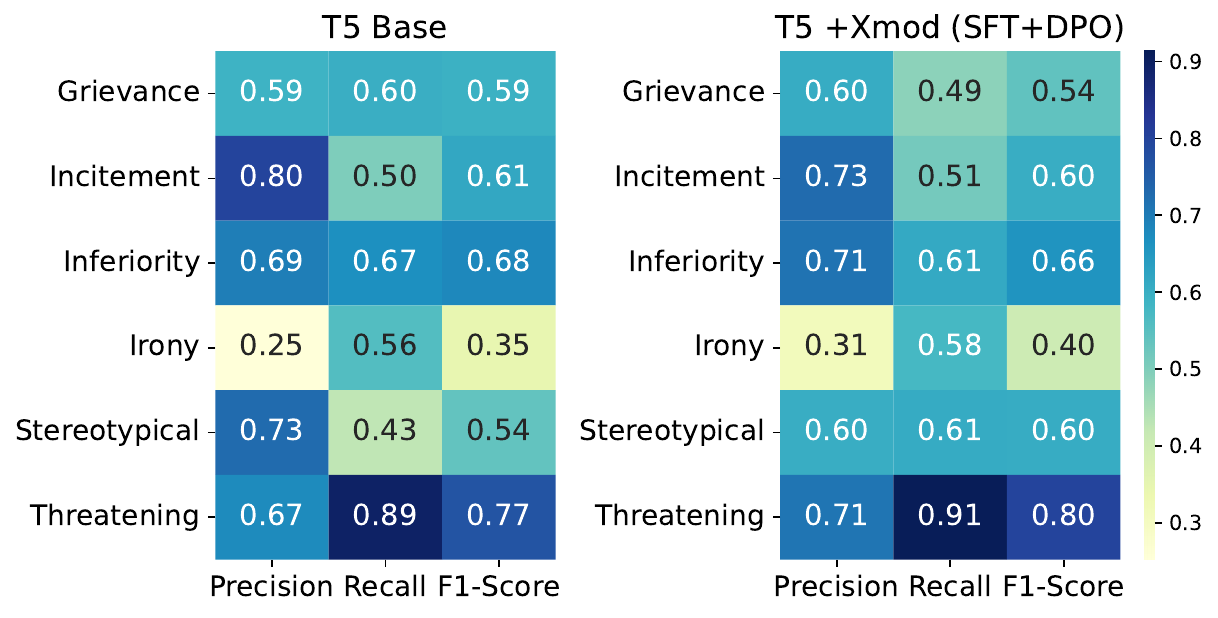}
        \includegraphics[width=0.48\linewidth]{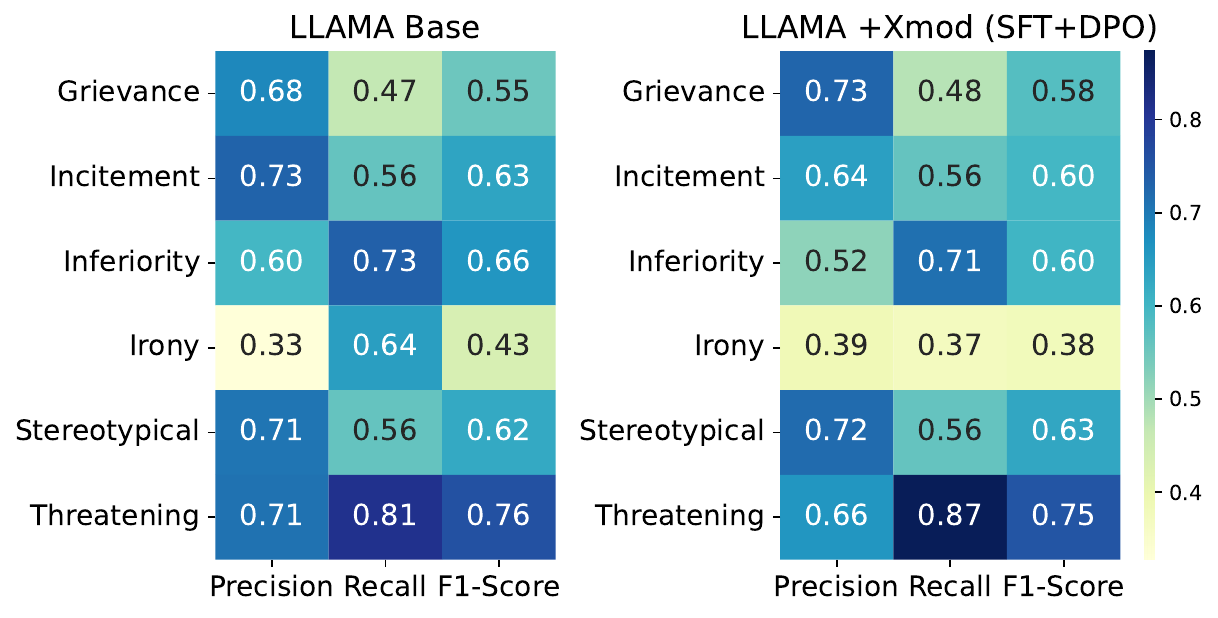}
        \caption{\textit{Implicit Hate}}
        \label{fig:latent_error}
    \end{subfigure}

    \caption{Comparison of Precision, Recall, and Macro F1-scores for the \textit{T5} and \textit{Llama} model families under two training regimes: self-augmentation only (\textit{Base}) and cross-model refinement (\textit{+Xmod}) at $K=256$. Models trained with cross-model refinement (right plot within each model family) generally show improvements in label categories where their counterparts already perform well, and conversely exhibit declines in areas corresponding to their weaknesses. }
    \label{fig:xmodel_anl}
\end{figure*}

\section{Qualitative Analysis of Explanation}
\label{apx:nli_edge}
We examine the relatively small amount of non-Entail samples in Section \ref{sec:nli_eval} to observe general trends in the models' inconsistency with respect to both the predicted label and definition.

First, \textit{T5 models typically exhibit higher consistency in explanation compared to their Llama counterparts} as shown in \autoref{tab:nli}. Even after cross-model refinement (XMOD), the percentage of Contradiction increases by at most 1\% (\textit{Implicit Hate}).  While a conclusive explanation is out of scope, we hypothesize that \textit{T5}'s encoder-decoder architecture supports higher robustness compared to \textit{Llama}'s decoder-only. 

Second, qualitative analysis reveals different patterns of inconsistency in the explanations offered by these 2 models. For \textit{HateXplain} and \textit{Implicit Hate}, \textit{Llama}'s Contradict-tagged explanations tend to consistently mentions some negative and/or offensive sentiments in the post, yet ultimately yielding the conflicting label \textit{Normal} (\autoref{fig:nli_spotcheck}).  In contrast, \textit{T5} Contradict explanations -- much fewer in overall number -- generally asserts a label at the beginning yet explicitly rejects this label at the end. For the more challenging \textit{Implicit Hate}, both models' Contradict explanations 
typically contain "lost in context" inconsistency, where the justification is grounded in some other labels' definition rather than the stated one, possibly due to the nuanced complexity of these finer-grained categories.

\paragraph{Our NLI-based consistency check also emerges as a simple yet useful quality check for downstream deployment.} In addition to classification metrics,  moderators may opt to implement a secondary lightweight NLI-checker to alert when a threshold of inconsistency is reached.

\section{Technical Details}
\label{apx:implement}
\subsection{Hardware} All experiments are carried out on a maximum of 2 Nvidia
RTX A6000 GPUs. We download the models from their Huggingface\footnote{\url{https://huggingface.co/}} checkpoints. Finetuning is implemented with LoRA techniques \cite{hulora} via the TRL library \footnote{\url{https://huggingface.co/docs/trl/en/index}}. The LoRA configurations for all settings are: 
\begin{itemize}
    \item Rank: 64
    \item \textit{alpha}: 128
    \item Dropout: 0.05
    \item Target modules: $q$ and $v$ in projection layers
\end{itemize}

\subsection{Inference Parameters}
All models in our experiments use the following parameters during inference:
\begin{itemize}
    \item temperature: 0.0
    \item max\_tokens/max\_new\_tokens:
        \begin{itemize}
            \item 512 for classification with explanation
            \item 20 for classification without explanation
        \end{itemize}
\end{itemize}

The library vLLM\footnote{\url{https://docs.vllm.ai/en/latest/}} is used coupled with the openAI \texttt{v1/chatcompletion} endpoint to enable fast inference in our experiments. The OpenAI model \texttt{GPT-5-chat-latest} endpoint was accessed on September 12, 2025. We also use the \texttt{Qwen2.5-32B-Instruct-AWQ} version on HuggingFace.

\subsection{Preference Optimization}
\label{apx:alignment}

\paragraph{Direct Preference Optimization} DPO is an alignment technique used to fine-tune large language models (LLMs) to better align with human preferences. Unlike traditional methods such as Reinforcement Learning from Human Feedback (RLHF), DPO simplifies the process by directly optimizing the policy model using a binary cross-entropy objective, thereby eliminating the need for a separate reward model and the complexities of reinforcement learning. This approach reparameterizes the reward model such that its optimal policy can be expressed in a closed form, transforming preference optimization into a classification problem \cite{rafailov2024direct}.

\paragraph{Kahneman-Tversky Optimization} KTO is another model alignment technique that aims to directly maximize the utility of a model's generations by applying principles from Kahneman-Tversky prospect theory \cite{kahneman2013prospect}. KTO evaluates outputs based on their perceived gains or losses relative to a reference point, incorporating the concept of loss aversion. This method utilizes data that simply indicates whether an output is desirable or undesirable for a given input, which can be more readily available and less expensive to collect than the paired preference data required by DPO \cite{ethayarajh2024kto}. 

\paragraph{Implementation} We primarily use the implementation of the $\texttt{DPOTrainer}$ and $\texttt{KTOTrainer}$ classes via the TRL library hosted on the HuggingFace platform. For the $\texttt{DPOTrainer}$, we set $\beta=0.1$, and the loss to be default \textit{sigmoid} loss. For the $\texttt{KTOTrainer}$, we similarly set the $\beta=0.1$ and use the default $\texttt{KTO loss}$ option.

\subsection{Hyperparameter Tuning}
For ModernBERT, we use the HuggingFace Trainer and train for 4 epochs with AdamW optimizer. We use the associated tokenizer and pad the inputs to maximum length (512) for each batch. 

For Base SFT finetuning, we use 3 epochs and learning rate 3e-4. We present the  chosen parameters for DPO/KTO alignment tuning in \autoref{tab:hatexplain_params}, \autoref{tab:latenthate_params}, \autoref{tab:Implicit_params} below.

\begin{table*}[h!]
\centering

\begin{subtable}[]{\linewidth}
    \footnotesize
    \centering
        \begin{tabular}{c|c|cc|cc}
        \toprule
        \multicolumn{1}{c|}{} & \multicolumn{1}{c|}{} & \multicolumn{2}{c|}{\textbf{T5}} & \multicolumn{2}{c}{\textbf{Llama}} \\
        \textbf{K} & \textbf{Technique} & \textbf{Epochs} & \textbf{LR} & \textbf{Epochs} & \textbf{LR} \\
        \midrule
        16 & DPO & 3 & 5e-05 & 3 & 1e-05 \\
        32 & DPO & 3 & 5e-05 & 4 & 1e-05 \\
        64 & DPO & 3 & 1e-05 & 3 & 1e-05 \\
        128 & DPO & 3 & 5e-05 & 3 & 1e-05 \\
        256 & DPO & 4 & 1e-04 & 3 & 1e-05 \\
        256 & KTO & -- & -- & 3 & 5e-07 \\
        \bottomrule
        \end{tabular}
    \caption{Hyperparameters for regular training.}
\end{subtable}
\hfill
\begin{subtable}[]{\linewidth}
    \footnotesize
    \centering
        \begin{tabular}{c|c|cc|cc}
        \toprule
        \multicolumn{1}{c|}{} & \multicolumn{1}{c|}{} & \multicolumn{2}{c|}{\textbf{T5}} & \multicolumn{2}{c}{\textbf{Llama}} \\
        \textbf{K} & \textbf{Technique} & \textbf{Epochs} & \textbf{LR} & \textbf{Epochs} & \textbf{LR} \\
        \midrule
        256 & DPO-K128 & 3 & 1e-05 & 4 & 1e-04 \\
        256 & DPO-K192 & 3 & 1e-04 & 1 & 7e-05 \\
        256 & DPO-N128 & 3 & 5e-05 & 5 & 5e-06 \\
        256 & DPO-N192 & 3 & 5e-05 & 5 & 5e-06  \\

        \bottomrule
        \end{tabular}
    \caption{Hyperparameters for subsampling training.}
\end{subtable}
\caption{\textit{HateXplain}.}
\label{tab:hatexplain_params}
\end{table*}

\begin{table*}[h!]
\centering

\begin{subtable}[]{\linewidth}
    \footnotesize
    \centering
    \begin{tabular}{c|c|cc|cc}
    \toprule
    \multicolumn{1}{c|}{} & \multicolumn{1}{c|}{} & \multicolumn{2}{c|}{\textbf{T5}} & \multicolumn{2}{c}{\textbf{Llama}} \\
    \textbf{K} & \textbf{Technique} & \textbf{Epochs} & \textbf{LR} & \textbf{Epochs} & \textbf{LR} \\
    \midrule
    16 & DPO & 3 & 5e-05 & 3 & 1e-06 \\
    32 & DPO & 3 & 1e-04 & 3 & 1e-06 \\
    64 & DPO & 3 & 5e-05 & 3 & 1e-06 \\
    128 & DPO & 4 & 5e-05 & 3 & 1e-04 \\
    256 & DPO & 3 & 1e-04 & 3 & 1e-04 \\
    256 & KTO & -- & -- & 3 & 5e-07 \\
    \bottomrule
    \end{tabular}
    \caption{Hyperparameters for regular training.}
\end{subtable}
\hfill
\begin{subtable}[]{\linewidth}
    \footnotesize
    \centering
    \begin{tabular}{c|c|cc|cc}
    \toprule
    \multicolumn{1}{c|}{} & \multicolumn{1}{c|}{} & \multicolumn{2}{c|}{\textbf{T5}} & \multicolumn{2}{c}{\textbf{Llama}} \\
    \textbf{K} & \textbf{Technique} & \textbf{Epochs} & \textbf{LR} & \textbf{Epochs} & \textbf{LR} \\
    \midrule
    256 & DPO-K128 & 4 & 1e-04 & 4 & 1e-04 \\
    256 & DPO-K192 & 3 & 1e-04 & 3 & 1e-04 \\
    256 & DPO-N128 & 4 & 1e-04 & 3 & 1e-04 \\
    256 & DPO-N192 & 4 & 1e-04 & 5 & 1e-04 \\
    \bottomrule
    \end{tabular}
    \caption{Hyperparameters for subsampling training.}
\end{subtable}
\caption{\textit{Latent Hate}.}
\label{tab:latenthate_params}
\end{table*}

\begin{table*}[h!]
\centering
\begin{subtable}[]{\linewidth}
    \footnotesize
    \centering
    \begin{tabular}{c|c|cc|cc}
    \toprule
    \multicolumn{1}{c|}{} & \multicolumn{1}{c|}{} & \multicolumn{2}{c|}{\textbf{T5}} & \multicolumn{2}{c}{\textbf{Llama}} \\
    \textbf{K} & \textbf{Technique} & \textbf{Epochs} & \textbf{LR} & \textbf{Epochs} & \textbf{LR} \\
    \midrule
    16 & DPO & 3 & 5e-05 & 3 & 5e-06 \\
    32 & DPO & 3 & 5e-05 & 4 & 5e-05 \\
    64 & DPO & 3 & 1e-04 & 3 & 1e-06 \\
    128 & DPO & 1 & 7e-05 & 3 & 1e-04 \\
    256 & DPO & 1 & 5e-05 & 1 & 5e-05 \\
    256 & KTO & -- & -- & 3 & 5e-07 \\
    \bottomrule
    \end{tabular}
    \caption{Hyperparameters for regular training.}
\end{subtable}
\hfill
\begin{subtable}[]{\linewidth}
    \footnotesize
    \centering
    \begin{tabular}{c|c|cc|cc}
    \toprule
    \multicolumn{1}{c|}{} & \multicolumn{1}{c|}{} & \multicolumn{2}{c|}{\textbf{T5}} & \multicolumn{2}{c}{\textbf{Llama}} \\
    \textbf{K} & \textbf{Technique} & \textbf{Epochs} & \textbf{LR} & \textbf{Epochs} & \textbf{LR} \\
    \midrule
    256 & DPO-K128 & 1 & 7e-05 & 1 & 1e-05 \\
    256 & DPO-K192 & 1 & 7e-05 & 1 & 5e-05 \\
    256 & DPO-N128 & 1 & 7e-05 & 1 & 1e-05 \\
    256 & DPO-N192 & 1 & 7e-05 & 1 & 5e-05 \\
    \bottomrule
    \end{tabular}   
    \caption{Hyperparameters for subsampling training.}
\end{subtable}
\caption{\textit{Implicit Hate}}
\label{tab:Implicit_params}
\end{table*}

\begin{table*}[ht]
\centering
\begin{tabular}{|p{3cm}|p{12cm}|}
\hline
\textbf{Dataset} & \textbf{Definition} \\
\hline

HateXplain &
\begin{itemize}
    \item \textbf{Normal:} neither hate speech nor offensive.
    \item \textbf{Hate Speech:} language which attacks, demeans,
offends, threatens, or insults a group based on race, ethnic origin, religion, disability, gender, age, sexual orientation, or other traits. it is not the presence of certain words that makes the text hate speech, rather you should look the context the word is used in the text.
    \item \textbf{Offensive Language:} usage of rude, hurtful, derogatory, obscene or insulting language to upset or embarrass people.
\end{itemize} \\

\hline

Latent Hate &
\begin{itemize}
    \item \textbf{Not Hate:} speech or actions that do not involve any form of hatred, prejudice, or discrimination
toward individuals or groups based on their characteristics.
    \item \textbf{Explicit Hate:} openly expressed,
direct forms of hatred and prejudice toward individuals or groups based on their characteristics.
    \item \textbf{Implicit Hate:} coded or indirect language that disparages a person or group on the basis of protected characteristics like race, gender, and cultural identity.
\end{itemize} \\

\hline

Implicit Hate &
\begin{itemize}
    \item \textbf{White Grievance:} includes frustration over a minority groups' perceived privilege and casting
majority groups as the real victims of racism. This language is linked to extremist behavior and
support for violence.
    \item \textbf{Inferiority Language:} implies one group or
individual is inferior to another, and it can include dehumanization, denial of a person’s humanity, and toxic language that compares the target with disease, insects, animals. Related to assaults on human dignity, dominance, and declarations of superiority of the in group.
    \item \textbf{Irony:} refers to the use of sarcasm , humor, and satire to attack or demean a protected class or individual.
    \item \textbf{Stereotypes And Misinformation:} associate a protected class with negative attributes such as crime, or terrorism. includes misinformation that feeds stereotypes and vice versa, like Holocaust
denial and other forms of historical negationism.
    \item \textbf{Threatening And Intimidation:} conveys
a speaker’s commitment to a target’s pain, injury, damage, loss or violation of rights, threats related to implicit violation of rights and freedoms, removal of opportunities, and more subtle
forms of intimidation.
    \item \textbf{Incitement to Violence:} includes flaunting in group unity and power
or elevating known hate groups and ideologies
\end{itemize} \\

\hline
\end{tabular}
\caption{Definitions of toxic label categories and definitions (collected from their original publications) across datasets.}
\label{tab:hate_definitions}
\end{table*}

\section{Examples of Explanations}
We provide some examples of the explanations generated during our cross-training phase in \autoref{fig:hatexplain_exp}, \autoref{fig:latent_exp}, and \autoref{fig:implicit_exp}.

\clearpage
\definecolor{lightpink}{HTML}{FFB6C1}
\definecolor{lightblue}{HTML}{ADD8E6}

\begin{figure*}
    \centering

    \begin{subfigure}[t]{0.95\textwidth}
        \colorbox{lightblue}{%
            \parbox{\linewidth}{%
                \ttfamily
                \textbf{POST:} \\
                what kind of brzen hapless *** uses an expression like "there’s no big *** there" \\
                \textbf{LABEL:} HATE
            }
        }
        \caption{A sample post in \textit{HateXplain} with gold label. Offensive content partially redacted.}
    \end{subfigure}

    \begin{subfigure}[t]{0.95\textwidth}
        \colorbox{lightgreen}{%
            \parbox{\linewidth}{%
                \ttfamily
            \textbf{VARIANT:} T5 DPO @ K=128 \\
\textbf{EXPLANATION:} The post targets the group of homosexuals by using a derogatory term "***" to insult and demean them. This language is offensive as it is rude, hurtful, and insulting. Therefore, the post is labeled as Offensive. \\
\textbf{LABEL}: Offensive
            }
        }
\vspace{1em}

    \end{subfigure}
    \hfill
    \begin{subfigure}[t]{0.95\textwidth}
        \colorbox{lightgreen}{%
            \parbox{\linewidth}{%
                \ttfamily
            \textbf{VARIANT:} T5 +XMOD SFT+DPO @ K= 256 \\
\textbf{EXPLANATION:} The post targets the group of individuals who identify as "brzen" and use the derogatory term "***" The language used in the post attacks and demeans this group based on their sexual orientation, which falls under the definition of hate speech. \\
\textbf{LABEL:} Hate
            }
        }

    \end{subfigure}
\vspace{1em}

    \begin{subfigure}[t]{0.95\textwidth}
        \colorbox{lightpink}{%
            \parbox{\linewidth}{%
                \ttfamily
            \textbf{VARIANT:} LLAMA DPO @ K=128 \\
\textbf{EXPLANATION:} The post targets the homosexual community by using a derogatory slur "***" to demean and insult them. This language is offensive as it is rude, hurtful, and insulting towards individuals based on their sexual orientation. Therefore, the post is labeled as Offensive. \\
\textbf{LABEL:} Offensive
            }
        }
    \end{subfigure}
\vspace{1em}

    \begin{subfigure}[t]{0.95\textwidth}
        \colorbox{lightpink}{%
            \parbox{\linewidth}{%
                \ttfamily
            \textbf{VARIANT:} LLAMA +XMOD SFT+DPO @ K=256 \\
\textbf{EXPLANATION:} The post targets the homosexual community by using a derogatory term "***" to insult and demean them. This language attacks and demeans a specific group based on their sexual orientation, which falls under the definition of hate speech. Therefore, the post is labeled as Hate. \\
\textbf{LABEL}: Hate
            }
        }
    \end{subfigure}
    \caption{Examples of a post in \textit{HateXplain} with gold label, along with the explanations of the \textit{T5} and \textit{Llama} self-augmented variants at K=128, and the cross-trained model using all K=256 in total. We observe that models trained with more data (+Xmod) gives the correct classification. }
    \label{fig:hatexplain_exp}
\end{figure*}

\begin{figure*}
    \centering

    \begin{subfigure}[t]{0.95\textwidth}
        \colorbox{lightblue}{%
            \parbox{\linewidth}{%
                \ttfamily
                \textbf{POST:} \\
it means people aren't buying the lie! diversity k**s  now prove me wrong \\
                \textbf{LABEL:} IMPLICIT HATE
            }
        }
        \caption{A sample post in \textit{Latent Hate} with gold label. Offensive content partially redacted.}
    \end{subfigure}

    \begin{subfigure}[t]{0.95\textwidth}
        \colorbox{lightgreen}{%
            \parbox{\linewidth}{%
                \ttfamily
            \textbf{VARIANT:} T5 DPO @ K=128 \\
\textbf{EXPLANATION:}  This post does not involve any form of hatred, prejudice, or discrimination towards individuals or groups based on their characteristics. It is a statement expressing skepticism towards the belief that diversity ***, but it does not disparage or express any form of hate towards any specific person or group. Therefore, it can be classified as "Not Hate." \\
\textbf{LABEL}: Not Hate
            }
        }
\vspace{1em}

    \end{subfigure}
    \hfill
    \begin{subfigure}[t]{0.95\textwidth}
        \colorbox{lightgreen}{%
            \parbox{\linewidth}{%
                \ttfamily
            \textbf{VARIANT:} T5 +XMOD SFT+DPO @ K= 256 \\
\textbf{EXPLANATION:} This post does not involve any form of hatred, prejudice, or discrimination towards individuals or groups based on their characteristics. The post simply states that people are not buying the lie that diversity ***, and asks for proof to prove the opposite. There is no indication of any negative or discriminatory language towards any specific group or entity. Therefore, this post should be classified as "Not Hate."  \\
\textbf{LABEL:} Not Hate
            }
        }

    \end{subfigure}
\vspace{1em}

    \begin{subfigure}[t]{0.95\textwidth}
        \colorbox{lightpink}{%
            \parbox{\linewidth}{%
                \ttfamily 
            \textbf{VARIANT:} LLAMA DPO @ K=128 \\
\textbf{EXPLANATION:} The post uses coded or indirect language to disparage a person or group based on their race.  Overall, the use of the phrase "diversity ***" and the challenge to "prove me wrong" indicate a negative view of diversity, suggesting a belief that diversity is harmful. This coded language indirectly disparages people of color, making it an example of implicit hate. \\
\textbf{LABEL:} Implicit Hate
            }
        }
    \end{subfigure}
\vspace{1em}

    \begin{subfigure}[t]{0.95\textwidth}
        \colorbox{lightpink}{%
            \parbox{\linewidth}{%
                \ttfamily
            \textbf{VARIANT:} LLAMA +XMOD SFT+DPO @ K=256 \\
\textbf{EXPLANATION:} This post should be classified as "Not Hate" because it does not involve any form of hatred, prejudice, or discrimination towards individuals or groups based on their characteristics. The post simply expresses a belief that diversity is a lie and challenges the reader to prove the speaker wrong. \\
\textbf{LABEL}: Not Hate
            }
        }
    \end{subfigure}
    \caption{Examples of a post in \textit{Latent Hate} with gold label, along with the explanations of the \textit{T5} and \textit{Llama} self-augmented variants at K=128, and the cross-trained model using all K=256 in total. In this example, we observe that \textit{T5} absorbs the verbosity of \textit{Llama} after cross-training, yet still gets the wrong label. On the other hand, \textit{Llama}'s explanation is more terse after cross-training, and also classifies the post incorrectly.  }
    \label{fig:latent_exp}
\end{figure*}

\clearpage
\begin{figure*}
    \centering

    \begin{subfigure}[t]{0.95\textwidth}
        \colorbox{lightblue}{%
            \parbox{\linewidth}{%
                \ttfamily
                \textbf{POST:} hi can i get a couple of grammes of your *** power ? \\
 
                \textbf{LABEL:} INCITEMENT
            }
        }
        \caption{A sample post in \textit{Implicit Hate} with gold label. Offensive content partially redacted.}
    \end{subfigure}
\hfill
    \begin{subfigure}[t]{0.95\textwidth}
        \colorbox{lightgreen}{%
            \parbox{\linewidth}{%
                \ttfamily
            \textbf{VARIANT:} T5 DPO @ K=128 \\
\textbf{EXPLANATION:} The post targets minorities by expressing frustration over their perceived privilege and casting the majority group (*** people) as the real victims of racism. The post uses language that elevates *** power and unity, which is a known hate group ideology. The post also includes a sarcastic and ironic tone by asking for a "couple of grammes of your *** power," implying that minorities are inferior and need to be subjugated. This aligns with the definition of incitement, as it flaunts in-group unity and power while also elevating known hate groups and ideologies.  \\
\textbf{LABEL}: Incitement
            }
        }
    \end{subfigure}
\vspace{1em}

    \begin{subfigure}[t]{0.95\textwidth}
        \colorbox{lightgreen}{%
            \parbox{\linewidth}{%
                \ttfamily
            \textbf{VARIANT:} T5 +XMOD SFT+DPO @ K= 256 \\
\textbf{EXPLANATION:}  The post flaunts in-group unity and power by flaunting the idea of *** power. By using the term "*** power," the post elevates the idea of *** power and elevates the hate group ideology of *** supremacy. This aligns with the definition of incitement, which includes flaunting in-group unity and power or elevating known hate groups and ideologies.  \\
\textbf{LABEL:} Incitement
        }
}
    \end{subfigure}
\vspace{1em}

    \begin{subfigure}[t]{0.95\textwidth}
        \colorbox{lightpink}{%
            \parbox{\linewidth}{%
                \ttfamily 
            \textbf{VARIANT:} LLAMA DPO @ K=128 \\
\textbf{EXPLANATION:} This post targets *** people and their perceived desire for power and authority. By specifically asking for a couple of grammes of *** power, the post elevates and promotes the idea of *** supremacy, which is a hate group and ideology that advocates for the superiority of the *** race. This incitement of in-group unity and power aligns with the definition of incitement, as it promotes and supports a hate group and their extremist behavior. \\
\textbf{LABEL:} Incitement
            }
        }
    \end{subfigure}
\vspace{1em}

    \begin{subfigure}[t]{0.95\textwidth}
        \colorbox{lightpink}{%
            \parbox{\linewidth}{%
                \ttfamily
            \textbf{VARIANT:} LLAMA +XMOD SFT+DPO @ K=256 \\
\textbf{EXPLANATION:}
            The post directly flaunts in-group unity and power by referencing *** power, which is a known hate group and ideology. The post specifically asks for a couple of grammes of *** power, which further promotes the idea of *** supremacy. This aligns with the definition of incitement, as it flaunts in-group unity and power and elevates known hate groups and ideologies. \\
\textbf{LABEL}: Incitement
        }
    }
    \end{subfigure}
    \caption{ Examples of a post in \textit{Implicit Hate} with gold label, along with the explanations of the \textit{T5} and \textit{Llama} self-augmented variants at K=128, and the cross-trained model using all K=256 in total. In this example, we also observe the cross-pollination of style in terms of verbosity, albeit to a lesser degree, but all models classify correctly.  }
    \label{fig:implicit_exp}
\end{figure*}

\begin{figure*}
    \centering

    \begin{subfigure}[t]{0.95\textwidth}
        \colorbox{lightblue}{%
            \parbox{\linewidth}{%
                \ttfamily
                \textbf{LLAMA - HATEXPLAIN} \\[0.4em]
            EXPLANATION: The post targets the group of homosexuals by using the term "queer" in a derogatory manner. The language used attacks and demeans this group based on their sexual orientation, which falls under the definition of hate speech. Therefore, the post is labeled as "Normal." \par LABEL: Normal
                \vspace{2.2em} 
            }
        }
        \caption{Llama: The final label `Normal' does not follow from the prior justification.}
    \end{subfigure}

    \vspace{1em}

    \begin{subfigure}[t]{0.95\textwidth}
        \colorbox{lightgreen}{%
            \parbox{\linewidth}{%
                \ttfamily
                \textbf{T5 - LATENT HATE} \\[0.4em]
               EXPLANATION: This post should be classified as "Not Hate" because it does not involve any form of hatred, prejudice, or discrimination towards individuals or groups based on their characteristics. The post simply mentions Alex Linder and white nationalism 1.0, without expressing any negative or discriminatory views towards any specific group or individual. Therefore, it does not meet the criteria for being classified as "Not Hate." \par LABEL: Not Hate
                \vspace{2.2em} 
            }
        }
        \caption{T5: The explanation starts with an asserted label yet later refutes this category.}
    \end{subfigure}

    \vspace{1em}

    \begin{subfigure}[t]{0.95\textwidth}
        \colorbox{lightpink}{%
            \parbox{\linewidth}{%
                \ttfamily
                \textbf{LLAMA - IMPLICIT HATE} \\[0.4em]
            EXPLANATION: The post uses irony to attack and demean the idea of a wall, sarcastically suggesting that it should also prevent merpeople from the ocean. The use of humor and satire is used to mock the idea of a wall being a barrier to prevent all forms of transportation, including boats, tunnels, and airplanes, as well as the ocean. \par
            LABEL: Threatening
                \vspace{2.2em} 
            }
        }
        \caption{Llama: The explanation discusses factors directly relevant to `Irony', yet the predicted label `Threatening' is unrelated. }
    \end{subfigure}

    \caption{Examples of explanations that are judged to contains some forms of inconsistency with respect to the predicted label and/or its definition in our NLI-based analysis.}
    \label{fig:nli_spotcheck}
\end{figure*}

\end{document}